\documentclass{article}

\usepackage[main,final]{neurips_2026}

\usepackage[utf8]{inputenc} % allow utf-8 input
\usepackage[T1]{fontenc}    % use 8-bit T1 fonts
\usepackage{hyperref}       % hyperlinks
\usepackage{url}            % simple URL typesetting
\usepackage{booktabs}       % professional-quality tables
\usepackage{amsfonts}       % blackboard math symbols
\usepackage{nicefrac}       % compact symbols for 1/2, etc.
\usepackage{microtype}      % microtypography
\usepackage[table]{xcolor}  % colors (table option for \rowcolor)
\usepackage{graphicx}
\usepackage{amsmath}
\usepackage{multicol}
\usepackage{multirow}
\usepackage{makecell}
\usepackage{enumitem}
\usepackage{caption}
\usepackage{subfigure}
\usepackage{graphicx}

\title{Trading Sensing for Structure: Sparse IMU-EMG Fingertip Force Estimation via Neuro-inspired Structured Modeling}

\author{%
  Yang Gao \\
  South China University \\
  of Technology \\
  Guangzhou, China \\
  {\small\texttt{gaoyang2025@scut.edu.cn}}
  \And
  Yingjing Xiao \\
  East China Normal University \\
  Shanghai, China \\
  {\small\texttt{51275901140@stu.ecnu.edu.cn}}
  \And
  Junbin Ren \\
  East China Normal University \\
  Shanghai, China \\
  {\small\texttt{renjunbin001@gmail.com}}
  \AND
  Chenxu Zhang \\
  South China University \\
  of Technology \\
  Guangzhou, China \\
  {\small\texttt{ftcxzhang@mail.scut.edu.cn}}
  \And
  Wenbo Zhang \\
  South China University \\
  of Technology \\
  Guangzhou, China \\
  {\small\texttt{202310193158@mail.scut.edu.cn}}
  \And
  Zhanpeng Jin\thanks{Corresponding author.} \\
  South China University \\
  of Technology \\
  Guangzhou, China \\
  {\small\texttt{zjin@scut.edu.cn}}
}

\begin{document}

% \author{Yingjing Xiao}
% \affiliation{%
%   \institution{East China Normal University}
%   \city{Shanghai}
% \email{51275901140@stu.ecnu.edu.cn}

% \author{Zhichao Huang}
% \affiliation{%
%   \institution{East China Normal University}
%   \city{Shanghai}
% \email{51275901117@stu.ecnu.edu.cn}

% \author{Yang Gao}
% \authornote{Yang Gao is the corresponding author}
% \affiliation{%
%   \institution{East China Normal University}
%   \city{Shanghai}
% \email{gaoyang@cs.ecnu.edu.cn}

\maketitle

\begin{abstract}
% We present a novel neuro-biomechanical structured model for estimating fine-grained finger-level contact forces during dexterous grasping. Unlike existing methods that often rely on dense sensor arrays or generic black-box architectures, our approach leverages sparse wearable inputs—a thumb-mounted inertial measurement unit (IMU) and a single-channel wrist electromyography (EMG) sensor—a configuration dictated by practical constraints. Inspired by human sensorimotor control, our model incorporates structural inductive biases through distinct processing pathways: a feedforward path integrating thumb kinematics and grasp intent via task-dependent modulation and sparse learned gating (mimicking synergies), and a feedback path refining predictions via EMG-driven scaling (mimicking local muscle tuning). This architecture, embodying what we term a Neuro-Biomechanical Structural Prior (NBSP), promotes data efficiency, interpretability, and generalization. Extensive evaluations on multi-user, multi-object datasets demonstrate that our sparse-input NBSP achieves performance competitive with baselines using significantly denser sensing, highlighting that informed structure can effectively compensate for sensing sparsity ("Less is More") and offering advantages in data-scarce and practical deployment scenarios.
We present a Neuro-inspired Structured Model (NiSM) for fine-grained finger-level contact force estimation during dexterous grasping from sparse wearable sensing. NiSM uses a thumb-mounted inertial measurement unit (IMU) and a single-channel wrist electromyography (EMG) sensor, while deriving grasp context from an IMU-based hand-shape latent. This work studies the trade-off between sensing density and structural inductive bias: when sensor coverage is limited, inferred grasp context, kinematic coupling, and muscle-effort modulation can provide useful constraints for force prediction. Using human sensorimotor control as an architectural analogy, NiSM organizes computation into complementary pathways: a context-conditioned feedforward pathway maps thumb kinematics to finger-specific estimates through sparse learned gating, while an EMG-conditioned modulation pathway provides effort-dependent scaling. By explicitly encoding these structural priors, NiSM reduces reliance on dense sensor arrays and large generic black-box models. Evaluations on multi-user, multi-object datasets show that NiSM achieves performance competitive with a dense-sensing reference where comparable dense inputs are available, while using substantially fewer parameters, and is more robust than generic models under limited-data settings. These results indicate that structural priors can partially recover force-relevant information under sparse sensing, offering a practical direction for wearable finger force estimation under real-world deployment constraints.
\end{abstract}


\section{Introduction}
\label{sec:introduction}
Estimating finger-level contact forces is essential for physically grounded hand interaction, enabling tasks such as dexterous manipulation, rehabilitation, and teleoperation. Despite substantial progress in hand pose estimation, force estimation remains less developed, largely because contact force depends on inter-finger coordination, muscle activation, and object interaction dynamics that are difficult to observe with wearable sensors of limited coverage.

Existing methods often rely on dense sensing, such as multi-channel EMG~\cite{rubin2022finger,seo2024posture, 9956792} or data gloves~\cite{austin2022fabricated, geng2022cnn, 8206575, 10342129}, which can be impractical for everyday deployment due to bulk, discomfort, or cost. Recent lightweight alternatives, including camera-based systems~\cite{taylor2022gelslim,ko2023vision, MAHDIKHANLOU2023109217, GAO202225}, reduce the need for dense wearable instrumentation, but often remain sensitive to visibility conditions or rely on generic black-box regression. These approaches provide useful sensing channels, yet they do not explicitly model the structured motor constraints that shape finger coordination and force generation.

This setting raises an inductive bias--information trade-off: when sensing density is reduced, what structure can be encoded in the model to recover useful force-relevant information? Biological sensorimotor systems provide a natural source of such structure, using task context, coordinated motor patterns, and muscle-dependent modulation to generate dexterous forces from limited and distributed signals~\cite{bhattasali2022neural}. Motivated by these principles, we study whether neuro-inspired structural inductive biases can improve sparse-input force estimation by constraining how kinematic, intent, and muscle-activation information are integrated.

In this work, we present a neuro-inspired structured model for finger force estimation using a thumb-mounted IMU, a single-channel wrist EMG sensor, and a grasp-context inferred from IMU-derived hand-shape latent. This configuration is highly constrained yet practical for wearable deployment. Our model organizes computation into complementary pathways inspired by functional components of human sensorimotor control. A context-modulated feedforward pathway maps thumb kinematics to finger-specific force estimates through sparse learned gating, while an EMG-conditioned modulation pathway provides effort-dependent scaling. This design explicitly encodes task context, kinematic coupling, and muscle-effort modulation as structural priors, in contrast to generic black-box fusion models.

We evaluate the proposed model on multi-user, multi-object force estimation datasets and compare it with sparse-input sequence models, black-box fusion models, and dense-sensing baselines. The results show that NiSM achieves performance competitive with dense-sensing baselines while using substantially fewer parameters, and exhibits stronger robustness than generic models under limited-data settings. Through ablation analysis, we further quantify the contribution of each structural component, highlighting how model structure can recover part of the information typically supplied by denser sensing.

Our main contributions are:
\begin{itemize}
    \item We propose NiSM, a neuro-inspired structured model that studies how three inductive biases can recover part of the force-relevant information lost to missing sensing channels in sparse IMU-EMG force estimation: inferred grasp context, sparse kinematic coupling, and EMG-conditioned effort scaling.
    \item We build and evaluate a diverse multi-user, multi-object dataset covering realistic grasp types and force levels, enabling systematic evaluation of sparse-sensing force estimation models.
    \item We show that NiSM achieves performance competitive with dense-sensing references using substantially fewer parameters, and ablations reveal how inferred grasp context, sparse kinematic coupling, and EMG-driven effort scaling contribute force-relevant information.
\end{itemize}

% \begin{figure}[!tbp]
% \centering
% \includegraphics[width=0.4\textwidth]{figures/hardware_preview.png}
% \vspace{-10pt}
% \caption{Ring-watch Hardware.}
% \label{fig:finger-wrist-rationale}
% \vspace{-15pt}
% \end{figure}

\vspace{-1em}
\section{Related Work}
\subsection{Hand Interaction Sensing via Vision}

Visual cues, such as fingernail color changes, deformation patterns, and skin displacement, have been widely explored as non-contact proxies for estimating fingertip pressure and interaction forces. Mascaro and Asada~\cite{1266642} reconstructed three-axis fingertip forces and finger postures from subnail blood volume variations. Chen et al.~\cite{Chen_Westling_Edin_van} further estimated force, torque, and surface curvature using visual features from the fingernail and surrounding skin. Fallahinia et al.~\cite{9772273} employed deep neural networks for real-time tactile force estimation from video inputs. These methods reduce the need for wearable force sensors, but their accuracy depends heavily on high-resolution visual observations and remains sensitive to lighting, occlusion, and hand-object visibility.

Other vision-based approaches estimate pressure or interaction forces directly from RGB or egocentric observations. PressureVision++~\cite{10484532} predicts fingertip pressure from a single RGB image, while EgoPressure~\cite{zhao_2024} introduces synchronized RGB-D data for hand pose and pressure estimation. Object-centric methods infer forces from observed object motion or physical simulation, including works by Ehsani et al.~\cite{9157658}, Li et al.~\cite{li2019estimating}, and Pham et al.~\cite{pham2015towards}. Although these approaches provide valuable external sensing channels, they can struggle with static objects, limited viewpoints, or cluttered interactions. More importantly, many rely on end-to-end visual regression rather than explicitly modeling the internal sensorimotor structure that links intent, kinematics, muscle activation, and force generation.

\subsection{Hand Interaction Sensing via Wearable Sensors}

\subsubsection{Single-Modality Approaches}

Wearable surface electromyography (sEMG) and inertial measurement units (IMUs) are commonly used for hand interaction sensing because they are compact and compatible with mobile deployment. sEMG captures neuromuscular activation and has been used for coarse gesture classification~\cite{10.1145/3442381.3449890, 10.3389/fnbot.2019.00042, 9861657, Salatiello2021ContinuousDO} and global grip force estimation~\cite{FIALKOFF2022103550, 10460104, MAO2023104088}. However, sEMG alone lacks explicit kinematic information and has limited observability for spatially resolved fingertip forces. IMUs, in contrast, are effective for tracking hand and finger motion~\cite{10.1145/3699741, liang2021dualring, 10.1145/3534587, 10763441}, but they primarily measure kinematic states and do not directly capture muscle effort or contact dynamics~\cite{s24196307}. Thus, each modality provides only partial information: sEMG reflects internal effort, while IMUs encode motion. This observability gap motivates multimodal sensing and structured integration.

\subsubsection{Multimodal Fusion Approaches}

Several studies combine IMU and sEMG signals for gesture recognition or force estimation. Multimodal hand gesture classification has been demonstrated with forearm-mounted sensors and deep Q-networks~\cite{s22249613}, as well as with multi-channel wrist-worn systems~\cite{8141959}. For force-related tasks, prior work has used high-density sEMG and IMU to estimate wrist angle and grasp force~\cite{10460104}, or to predict endpoint force during muscle contractions~\cite{9825705}. These methods show the value of complementary sensing, but often operate at the level of discrete gestures, global force, or endpoint force rather than fine-grained per-finger contact estimation.

Recent dense multimodal systems provide richer signals for detailed hand interaction modeling. PiMForce~\cite{NEURIPS2024_a01e69aa}, for example, integrates dense sensing and posture information to estimate spatially distributed hand pressure under free-form interactions. Such systems demonstrate the benefit of high sensing density, but their instrumentation can be cumbersome for everyday deployment. In addition, many fusion models rely on generic deep architectures, leaving open the question of how much sensing density can be reduced when task structure, kinematic coupling, and effort-dependent modulation are explicitly encoded. Our work addresses this question by studying sparse wearable inputs together with structural inductive biases.

\subsection{Neuro-inspired Models for Motor Control}

\begin{figure}[!bp]
    \centering    \includegraphics[width=0.98\textwidth,height=0.2\textheight]{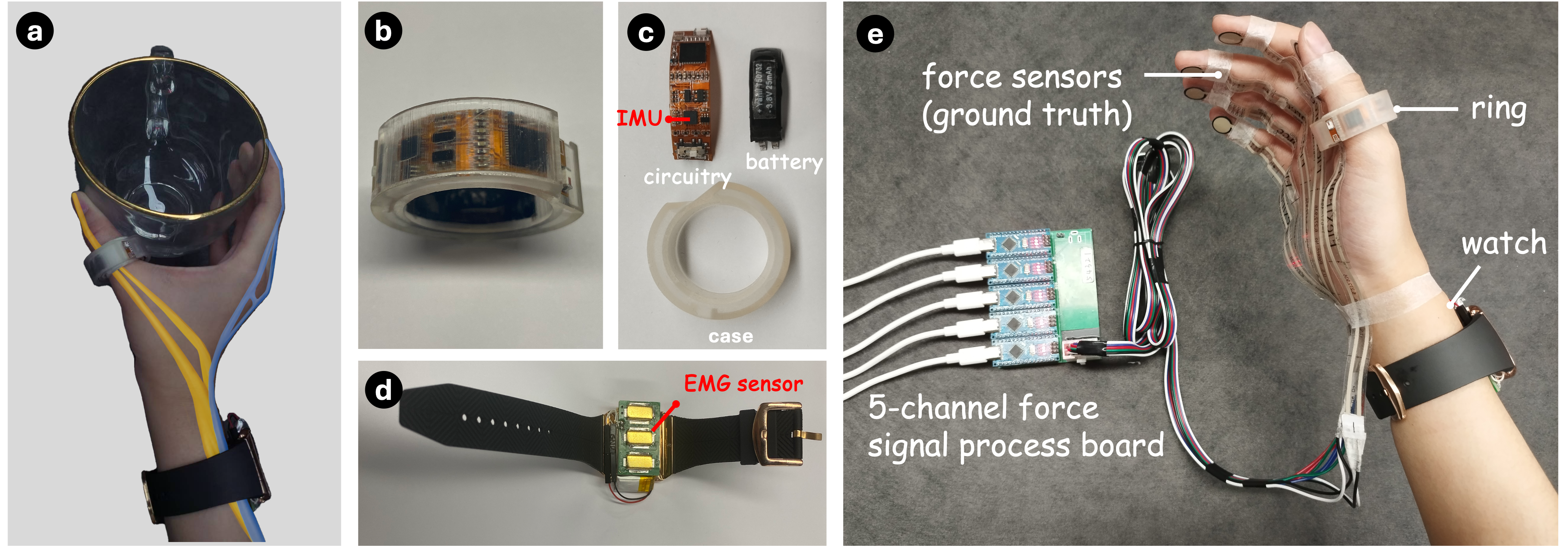}
    \caption{Minimalist Ring-Watch system for finger force sensing.(a) Hand interaction during grasping. (b–c) Custom ring with embedded IMU. (d) Watch with integrated EMG sensor. (e) Full setup with ground-truth force sensors and signal board.}
    \label{fig:hardware}
    \vspace{-15pt}
\end{figure}

Biological motor control has motivated many models for robot locomotion, arm movement, manipulation, and navigation. Examples include central pattern generator models for locomotion~\cite{billard2000biologically}, neural controllers for arm movement~\cite{bernabucci2007biologically}, and bio-inspired sensorimotor control frameworks using artificial or spiking neural networks~\cite{zahra2021differential,verduzco2022self,fernandez2021biological}. These studies show that structured computation can improve efficiency, adaptability, and robustness in embodied control.

However, most bio-inspired models target control generation rather than fine-grained per-finger force estimation from sparse wearable biosignals. They also often rely on dense sensory inputs, such as vision or detailed sensor arrays~\cite{zahra2021differential,guerrero2023bio}, which differs from the practical constraints of minimal wearable sensing. Although structured neural architectures have been studied as inductive priors for embodied control~\cite{bhattasali2022neural,verduzco2022self}, their use for recovering force-relevant information from sparse internal signals remains underexplored. In particular, task-dependent modulation, sparse synergy-like mappings, and EMG-conditioned effort scaling have not been systematically integrated for sparse finger force prediction.

To address this gap, NiSM uses the functional organization of sensorimotor control as an architectural template, combining context-conditioned feedforward prediction with EMG-conditioned scaling, and studies how this structural inductive bias can compensate for reduced sensing density in wearable force estimation.

\section{Methodology}

\subsection{Minimalist Ring-Watch Sensing System}
\label{sec:hardware}

To support force-aware hand interaction sensing in an everyday wearable form factor, we design a minimalist ring-watch sensing system consisting of a finger-worn ring with a 9-axis Inertial Measurement Unit (IMU) and a smartwatch-like device with a single-channel surface Electromyography (sEMG) sensor (Figure \ref{fig:hardware}). The thumb-mounted IMU captures local motion and orientation, while the wrist EMG records coarse muscle activation related to grasp effort. The ring is worn on the thumb because the thumb participates in nearly all grasp types and serves as the opposition reference for the other fingers. This sparse configuration balances sensing capability with comfort, portability, and social acceptability, mitigating the deployment limitations of bulkier data gloves or dense IMU/EMG arrays. The time-synchronized IMU and EMG streams provide complementary but partial observations—local thumb kinematics and global muscle-effort cues—which form the sparse sensing basis for our neuro-inspired structured model; task-level grasp intent is incorporated at the model level, as described in Section \ref{sec:NiSM}, to estimate per-finger contact pressure.

\subsection{Neuro-inspired Structured Model}
\label{sec:NiSM}

\begin{figure}[!tbp]
    \centering    \includegraphics[width=0.9\textwidth,height=0.28\textheight]{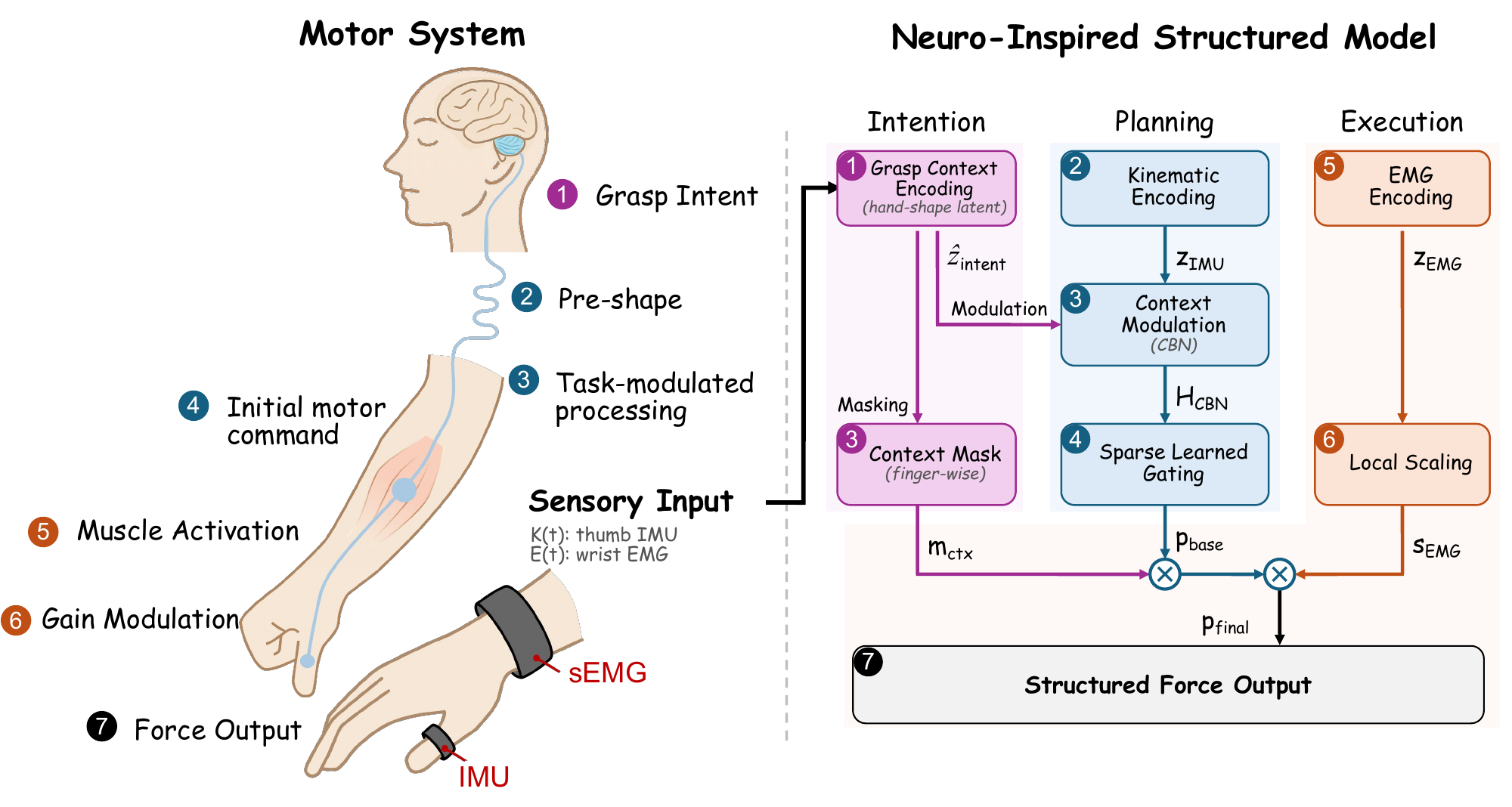}
    \caption{Overview of the Neuro-inspired Structured Model (NiSM) for Finger Force Prediction. Left: Biological Inspiration: A simplified schematic of the human motor system illustrating the hypothesized information flow from grasp intent (1) through sensorimotor processing (2-4) to muscle activation and gain modulation (5--6) and force output (7). Right: NiSM follows this organization with three stages: intention, planning, and execution. The thumb IMU stream $\mathbf{K}(t)$ is encoded by modules 1 and 2, and the wrist EMG $\mathbf{E}(t)$ by module 5. Module 1 infers the hand-shape latent $\hat{\mathbf{h}}$ and maps it to the context feature $\hat{\mathbf{z}}_{\mathrm{intent}}$ ($\hat{c}(t)$ in Eq.~\ref{eq:formulation}), which conditions the CBN modulation and the finger-wise context mask $\mathbf{m}_{\mathrm{ctx}}$.}
    \label{fig:model_overview}
    \vspace{-10pt}
\end{figure}

To study how structural inductive bias can compensate for reduced sensing density in wearable force estimation, we design NiSM, a neuro-inspired structured architecture for our minimalist \textit{ring-watch system} (Section \ref{sec:hardware}), as shown in Figure \ref{fig:model_overview}. Rather than relying on dense sensing or generic black-box fusion, NiSM encodes task context, kinematic coupling, and effort-dependent modulation as explicit structural priors.

The model predicts per-finger contact pressure from two sparse sensing channels — thumb kinematics from an IMU and forearm muscle activation from a single-channel EMG, with a grasp-context feature inferred from the IMU stream. These inputs are processed through interpretable modules inspired by human sensorimotor control, including context-conditioned kinematic processing, sparse finger-wise projection, and EMG-conditioned gain modulation.

We formulate sparse-input force estimation as:
\begin{equation}
    \mathbf{P}(t) = f\left(\mathbf{K}(t), \mathbf{E}(t)\right), \textrm{with}~\hat{c}(t) = g_{ctx}(\mathbf{K}(t))
    \label{eq:formulation}
\end{equation}
where $\mathbf{K}(t)$ represents the thumb IMU-derived kinematic state, $\mathbf{E}(t)$ denotes the synchronized EMG signal reflecting muscle-effort information, and the grasp-context feature $\hat{c}$ is inferred from $\mathbf{K}(t)$ via a hand-shape latent encoder $g_{ctx}$. In Figure~\ref{fig:model_overview} and below, $\hat{c}(t)$ is denoted $\hat{\mathbf{z}}_{\mathrm{intent}}$.

Given synchronized thumb IMU and EMG streams, NiSM outputs a five-dimensional vector $\mathbf{p}_{\text{final}} \in \mathbb{R}^5$ representing finger-level contact pressure estimates.

\subsubsection{Input Encoding}

This stage encodes sparse sensor measurements into compact feature representations for subsequent structured force prediction.

\paragraph{Thumb Kinematic Encoding} 
The thumb IMU sequence $\mathbf{K} \in \mathbb{R}^{T \times d_{\text{IMU}}}$, where $T=30$ and $d_{\text{IMU}}=12$ (3-axis acceleration + 9-element rotation matrix), is processed by an LSTM followed by a linear layer. This captures temporal dependencies in thumb motion and yields a fixed-size kinematic feature vector $\mathbf{z}_{\text{IMU}} \in \mathbb{R}^{d_z}$, providing motion context for inferring coupled finger-force patterns from sparse thumb observations.

% \paragraph{Grasp Intent Encoding} 
% A discrete grasp intent label $\mathbf{C} \in \{1, \dots, N_{\text{intent}}\}$ is first embedded into $\text{Emb}(c) \in \mathbb{R}^{d_{\text{emb}}}$ and then transformed by the intent encoder $E_{\text{Intent}}$, an MLP, into an intent feature vector $\mathbf{z}_{\text{Intent}} \in \mathbb{R}^{d_z}$. This feature provides task-context information that constrains the mapping from sparse kinematics to finger-specific force estimates.

\paragraph{Inferred Grasp-Context Encoding.}
NiSM does not directly embed ground-truth grasp-intent or object labels. Instead, it derives a continuous grasp-context feature from an IMU-based hand-shape latent. The thumb IMU sequence is encoded into a hand-shape representation \(\hat{\mathbf{h}}\), which is then mapped by an MLP context encoder into \(\hat{\mathbf{z}}_{\mathrm{intent}}\). Object/action labels, when available, are used only as auxiliary training supervision and are not provided to the model at inference. Both $\hat{\mathbf{z}}_{\mathrm{intent}}$ and $\mathbf{z}_{\text{IMU}}$ are derived from the thumb IMU, so $\hat{\mathbf{z}}_{\mathrm{intent}}$ serves as an auxiliary-supervised continuous task condition computed from the same observation.

\paragraph{EMG Activation Encoding}
Each single-channel wrist EMG window $\mathbf{E}_{\text{window}} \in \mathbb{R}^{T \times 1}$ is first converted into a compact 10-dimensional feature vector $\mathbf{f}_{\text{EMG}} \in \mathbb{R}^{10}$ via time- and frequency-domain feature extraction. This vector is then passed through an MLP-based encoder $E_{\text{EMG}}$, producing the muscle activation feature $\mathbf{z}_{\text{EMG}} \in \mathbb{R}^{d_z}$ after a final \texttt{tanh} activation, which serves as an effort-related feature for EMG-conditioned force modulation.

\subsubsection{Planning Pathway: Context-Modulated Command Synthesis}

This pathway implements an hand-shape-conditioned feedforward prediction from sparse thumb kinematics. Its design uses the inferred grasp context to condition how kinematic information is interpreted, and then applies sparse learned gating to produce finger-specific base pressure estimates.

\paragraph{Context-Conditioned Modulation:} 

The kinematic feature $\mathbf{z}_{\text{IMU}} \in \mathbb{R}^{d_z}$ is modulated by the context feature $\hat{\mathbf{z}}_{\mathrm{intent}}\ \in \mathbb{R}^{d_z}$ using Conditional BatchNorm (CBN)~\cite{de2017modulating}. The CBN layer dynamically adjusts its normalization parameters based on $\hat{\mathbf{z}}_{\mathrm{intent}}$:
\begin{equation}
    \mathbf{H}_{\text{CBN}} = \gamma(\hat{\mathbf{z}}_{\mathrm{intent}}) \odot \frac{\mathbf{z}_{\text{IMU}} - \mu_{\text{B}}}{\sigma_{\text{B}}} + \beta(\hat{\mathbf{z}}_{\mathrm{intent}})
    \label{eq:cbn}
\end{equation}
where $\mu_B$ and $\sigma_B$ are batch statistics of $\mathbf{z}_{\text{IMU}}$ over the current mini-batch, and $\odot$ denotes element-wise multiplication. The scaling parameters $\gamma(\hat{\mathbf{z}}_{\mathrm{intent}})$ $\in \mathbb{R}^{d_z}$ and shifting parameters $\beta(\hat{\mathbf{z}}_{\mathrm{intent}})$ $\in \mathbb{R}^{d_z}$ are affine transformations of $\hat{\mathbf{z}}_{\mathrm{intent}}$, learned by separate linear layers:
\begin{align}
    \gamma(\hat{\mathbf{z}}_{\mathrm{intent}}) &= \mathbf{W}_{\gamma}\hat{\mathbf{z}}_{\mathrm{intent}} + \mathbf{b}_{\gamma} \label{eq:gamma} \\
    \beta(\hat{\mathbf{z}}_{\mathrm{intent}}) &= \mathbf{W}_{\beta}\hat{\mathbf{z}}_{\mathrm{intent}} + \mathbf{b}_{\beta} \label{eq:beta} 
\end{align}
where $\mathbf{W}_{\gamma}, \mathbf{W}_{\beta} \in \mathbb{R}^{d_z \times d_z}$ and $\mathbf{b}_{\gamma}, \mathbf{b}_{\beta} \in \mathbb{R}^{d_z}$ are learnable parameters. This modulation encodes the inferred grasp context as a task condition, allowing the same sparse thumb motion to be interpreted differently under different grasp conditions and producing a task-contextualized feature $\mathbf{H}_{\text{CBN}} \in \mathbb{R}^{d_z}$.

\paragraph{Sparse Learned Gating:} 
The context-modulated kinematic feature $\mathbf{H}_{\text{CBN}}$ is projected to the base pressure prediction $\mathbf{p}_{\text{base}} \in \mathbb{R}^{N_f}$ through a learned sparse weight matrix $\mathbf{W}_{\text{sparse}} \in \mathbb{R}^{d_z \times N_f}$. Each column $\mathbf{W}_{\text{sparse},j} \in \mathbb{R}^{d_z}$ corresponds to the feature-to-pressure mapping for the $j$-th finger:
\begin{equation}
    p_{\text{base},j} = \mathbf{H}_{\text{CBN}}^\top \mathbf{W}_{\text{sparse},j}.
\end{equation}
The sparse projection acts as a structural constraint on the mapping from shared kinematic features to finger-specific pressure estimates. Its learned non-zero weights identify salient feature--finger dependencies, reflecting synergy-like coordination patterns while avoiding a fully dense black-box mapping. Sparsity in $\mathbf{W}_{\text{sparse}}$ is encouraged through $L_1$ regularization during training (Section \ref{sec:training_loss}).

\subsubsection{Execution Pathway: EMG-Conditioned Effort Modulation}

Complementary to the feedforward pathway, this module uses the EMG signal to provide effort-dependent, finger-wise modulation of the base pressure prediction. This pathway introduces muscle-activation information that is not directly observable from sparse thumb kinematics, serving as a structural prior for compensating reduced sensing density.

\paragraph{EMG-Conditioned Local Scaling:} 
In parallel with the feedforward pathway, the encoded EMG feature $\mathbf{z}_{\text{EMG}}$ is mapped to finger-wise gain factors. A linear layer followed by a sigmoid activation produces $\mathbf{s}_{\text{EMG}} \in [0,1]^5$, with one scaling factor for each finger. Rather than directly predicting pressure from EMG alone, this module uses muscle effort to modulate the base prediction generated from context-conditioned thumb kinematics. In this way, EMG provides an effort-related constraint that reweights each finger’s base pressure estimate, allowing NiSM to incorporate peripheral muscle-activation cues while avoiding a fully dense black-box fusion.

% The EMG feature $\mathbf{z}_{\text{EMG}}$ is transformed into a vector of scaling factors $\mathbf{s}_{\text{EMG}} \in [0, 1]^5$, one for each finger, using a linear layer and a sigmoid activation:
%     $$ \mathbf{s}_{\text{EMG}} = \sigma(\text{Linear}(\mathbf{z}_{\text{EMG}})) $$
%     Unlike simply concatenating EMG features, this design uses the EMG signal, a proxy for overall muscle effort, to generate **sparse, finger-specific gain coefficients**. This conceptually mirrors how local feedback mechaNiSMs might non-uniformly amplify or attenuate global motor commands for individual muscles or finger groups based on their current activation state or load.

\subsubsection{Structured Force Output}

The final finger pressure prediction $\mathbf{p}_{\text{final}}$ is obtained by multiplicatively combining the feedforward base prediction $\mathbf{p}_{\text{base}}$, the inferred context-dependent finger mask \(\mathbf{m}_{\mathrm{ctx}}\), and the EMG-conditioned scaling factors $\mathbf{s}_{\text{EMG}}$:
\begin{equation}
    \mathbf{p}_{\text{final}} = \text{ReLU}(\mathbf{p}_{\text{base}} \odot \mathbf{m}_{\mathrm{ctx}} \odot \mathbf{s}_{\text{EMG}})
\end{equation}
where $\odot$ denotes element-wise multiplication. \(\mathbf{m}_{\mathrm{ctx}} = M_{\mathrm{ctx}}(\hat{\mathbf{z}}_{\mathrm{intent}})\) provides a finger-wise task prior inferred from IMU-derived hand-shape context, while \(\mathbf{s}_{\mathrm{EMG}}\) introduces effort-dependent modulation from muscle activation. This structured interaction combines task, motion, and effort cues to constrain pressure prediction under sparse sensing, and the ReLU activation enforces non-negative outputs.

\subsection{Training Objective with Structural Regularization} 
\label{sec:training_loss}
The model parameters are optimized with a loss function that balances pressure prediction accuracy and structural regularization, encouraging NiSM to learn an information-efficient mapping under sparse sensing.

The primary objective is to minimize the Mean Squared Error (MSE) between the predicted pressures $\mathbf{p}_{\text{final}}$ and the ground-truth pressures $\mathbf{p}_{\text{GT}}$:
\begin{equation}
    L_{\text{MSE}} = ||\mathbf{p}_{\text{final}} - \mathbf{p}_{\text{GT}}||_2^2
\end{equation}

To promote sparsity in the learned gating matrix $\mathbf{W}_{\text{sparse}}$ within the feedforward pathway, we add an $\ell_1$ regularization term:
\begin{equation}
    L_{\text{sparse}} = ||\mathbf{W}_{\text{sparse}}||_1
\end{equation}
This encourages many weights in $\mathbf{W}_{\text{sparse}}$ to approach zero, forcing the model to rely on the most relevant context-modulated thumb kinematic features for each finger's base pressure prediction. This $\ell_1$ regularization acts as a structural prior, guiding the model toward a sparse and interpretable feature-to-finger mapping rather than an unconstrained dense projection. Additionally, to improve the reconstruction of critical peak pressure magnitudes, we incorporate a peak-pressure error term $L_{\text{peak}}$. This term penalizes discrepancies between predicted and ground-truth peak pressures, defined as the sum of squared differences between the maximum values of $\mathbf{p}_{\text{final},i}$ and $\mathbf{p}_{\text{GT},i}$ for each finger $i$ within the training window.

The total loss function is a weighted sum:
\begin{equation}
    L_{\text{total}} = L_{\text{MSE}} + \lambda_{\text{sparse}} L_{\text{sparse}} + \lambda_{\text{peak}} L_{\text{peak}} + \lambda_{\mathrm{shape}}\mathcal{L}_{\mathrm{shape}} +\lambda_{\mathrm{ctx}}\mathcal{L}_{\mathrm{ctx}}
\end{equation}
where \(\lambda_{\text{sparse}}\), \(\lambda_{\text{peak}}\), 
\(\lambda_{\mathrm{shape}}\), and \(\lambda_{\mathrm{ctx}}\) control sparsity regularization, peak-pressure reconstruction, auxiliary hand-shape supervision, and auxiliary context regularization, respectively. 
The last two terms are used only during training and do not require hand-shape targets or object/action labels at inference. 
Minimizing \(L_{\text{total}}\) promotes accurate pressure estimation while maintaining sparse structural bias under reduced sensing density.

\section{Experiments}
\label{sec:experiments}
\subsection{Dataset}

\paragraph{Self-Collected Dataset}
\label{sec:self_collected_data}
We constructed a dataset for finger-level contact pressure estimation under sparse wearable sensing conditions with IRB approval. In this work, fingertip contact pressure is used as a proxy for finger-level normal contact force. The dataset contains recordings from 20 participants performing 20 diverse hand gestures and grasp actions (e.g., power grasp, pinch, tool use), each repeated five times. Participants were encouraged to freely vary their force during repetitions, resulting in high intra-action variability and more natural grasping behavior. They also put on the ring and watch themselves without a mounting fixture, so device placements reflect natural everyday wear. The dataset comprises over 1 million time-aligned frames. Ground-truth fingertip normal-force readings were recorded at 100 Hz using FlexiForce sensors on all five fingers, with measurements up to 25 N, and were normalized as contact-pressure labels. These labels were synchronized with a thumb-mounted IMU and a single-channel wrist EMG using shared timestamps. Additionally, 3D hand poses were extracted for auxiliary analysis.

This dataset provides a sparse-sensing testbed for evaluating how thumb kinematics and muscle-effort cues complement each other under realistic grasp variability. We split it by participant into 16 training and 4 test users. A quantitative comparison with public benchmark data is provided in the Appendix.

\paragraph{Public Benchmark Dataset}
To evaluate NiSM under a densely instrumented benchmark setting, we use the PiMForce dataset~\cite{NEURIPS2024_a01e69aa}, a large-scale multimodal corpus for whole-hand pressure estimation. It includes 21 participants performing predefined grasp and manipulation tasks, recorded with 3D hand posture, 8-channel sEMG, and high-resolution pressure measurements covering fingertips and palm patches, totaling over 80 million multimodal frames.

Unlike our sparse ring-watch setup, PiMForce provides dense sensing and structured interaction protocols. We hold out 4 of its 21 participants for testing (Appendix~\ref{app:pimforce}). For fair comparison, we project both predicted and ground-truth outputs to a unified five-finger pressure space. This enables a comparison in a common output space between sparse structured modeling and dense-sensing baselines, allowing us to examine how much force-relevant information can be recovered under reduced sensing density.

% \subsection{Evaluation Metrics}

% To assess the accuracy and temporal consistency of predicted finger forces, we adopt four complementary metrics:
% \paragraph{R² (Coefficient of Determination)} Measures how well the predicted force values explain the variance in the ground-truth signals. Higher values indicate better overall fit.

% \paragraph{NRMSE (Normalized Root Mean Squared Error)} Captures the scale-normalized magnitude of prediction errors, enabling fair comparisons across fingers with different force ranges.

% \paragraph{MAE (Mean Absolute Error)} Quantifies the average absolute difference between predicted and true forces. It reflects how close predictions are to the ground truth in physical terms (e.g., Newtons).

% \paragraph{Pearson Correlation Coefficient} Evaluates the temporal correlation between predicted and ground-truth force sequences, focusing on trend alignment rather than magnitude.

\subsection{Baselines}
To evaluate NiSM under the sensing density--structural inductive bias trade-off, we compare it with baselines that differ in model structure and input density. These include generic time-series models trained on the same sparse inputs, a sparse-input black-box fusion model, and a dense-sensing reference model. Together, they assess NiSM's prediction performance, data efficiency, and the contribution of its structural priors under reduced sensing density.

\paragraph{Time-Series Models.} 
To compare NiSM with generic sequence modeling approaches, we adapt representative time-series models to the same sparse IMU and EMG inputs. These include:
\textbf{Informer~\cite{zhou2021informer}}, a Transformer-based encoder-decoder model for long-range temporal dependency modeling, configured for concurrent finger pressure prediction;
\textbf{LSTM-Transformer~\cite{pentsos2025hybrid}}, a hybrid recurrent-attention model adapted for per-frame pressure regression from sparse inputs; and
\textbf{Crossformer~\cite{zhang2023crossformer}}, a Transformer-based model that explicitly captures cross-dimension dependencies in multivariate time series, adapted to our multimodal pressure estimation task.
These baselines test whether generic temporal modeling alone can extract force-relevant information from limited wearable signals.

\paragraph{Sparse Input Black-Box Deep Learning Model.} 
This baseline evaluates whether a more complex generic fusion architecture can achieve comparable performance without NiSM's structural inductive biases. We include a generic encoder-decoder with Cross-Attention Transformer (CA-T)~\cite{10.1145/3706599.3720220}, which jointly predicts finger pressure and 3D hand posture from sparse IMU/EMG inputs. Although CA-T can exploit correlations between pose and pressure, it does not explicitly encode task-dependent modulation, sparse feature-to-finger mapping, or EMG-conditioned effort modulation, serving as a black-box sparse-input baseline.

\paragraph{Dense Input Reference Model.} 
This baseline provides a data-rich reference using denser sensing than our ring-watch setup. Comparing NiSM with this model allows us to examine whether structural inductive biases can recover part of the force-relevant information typically supplied by denser sensing. We implement and retrain the core encoder-decoder architecture from PiMForce~\cite{NEURIPS2024_a01e69aa}, a state-of-the-art model for hand pressure estimation. This baseline is trained with dense sensing modalities available in PiMForce, such as multi-channel EMG and 3D hand posture, while pressure measurements are used as supervision. For fair comparison, its predictions are mapped to the same unified five-finger pressure space.

\section{Results: How Much Force-Relevant Information Can Structure Recover?}
We organize the results around a central question: how much force-relevant information can neuro-inspired structure recover under reduced sensing density? 
We examine this through three analyses: comparison with sparse black-box models and a dense-sensing reference, evaluation under limited user data, and component ablations that reveal which priors recover missing force-relevant information.

\subsection{Can Structured Sparse Sensing Narrow the Gap to Dense Sensing?}
\label{sec:dense_gap}

We benchmark NiSM against representative models on both our self-collected dataset and the public PiMForce dataset. As shown in Table \ref{tab:performance_benchmark}, NiSM achieves the best overall performance among the evaluated baselines while using significantly fewer parameters (1.71M). On the PiMForce benchmark, where a dense-sensing reference is available, NiSM achieves R² = 54.40\% and PCC = 76.60\%, slightly outperforming the dense PiMForce model (R² = 52.68\%, PCC = 73.73\%), which relies on denser inputs and substantially more parameters (66.39M). The standard deviations reported in Table \ref{tab:performance_benchmark} are computed over the four held-out test users, each evaluated separately, and thus reflect inherent inter-user variability. Because this test set is small, we also run 21-fold leave-one-subject-out cross-validation on PiMForce (in each fold, one participant is held out for testing and another for checkpoint selection), which gives R\textsuperscript{2} = 55.04 $\pm$ 10.24\% and PCC = 76.23 $\pm$ 5.70\%, close to the fixed-split result. Paired comparisons over the same 21 folds show that NiSM improves R\textsuperscript{2} by 0.300 (95\% CI [0.226, 0.374]) over CA-T and by 0.167 ([0.128, 0.206]) over PiMForce (dense), with two-sided Wilcoxon signed-rank $p < 0.001$ for both R\textsuperscript{2} and PCC (Appendix~\ref{app:stats}).

This result highlights the effectiveness of NiSM's structured inductive biases, including task-context modulation, sparse feature-to-finger mapping, and EMG-conditioned effort modulation, suggesting that explicit structure can recover part of the force-relevant information otherwise provided by denser sensing. Since PiMForce (dense) differs from NiSM in input modalities and architecture, and all models are retrained on PiMForce, this comparison evaluates the architectures under the same sparse-view setting and does not evaluate cross-dataset or cross-hardware generalization.

\begin{table*}[htbp]
\centering
\scriptsize
\renewcommand{\arraystretch}{1.3}
\setlength{\tabcolsep}{1.6pt}
\begin{tabular}{ll|cccc|cccc}
\hline
\multicolumn{2}{c|}{\textbf{Across Users}} & \multicolumn{4}{c|}{\textbf{Self-collected}} & \multicolumn{4}{c}{\textbf{PiMForce}} \\
\textbf{Models} & \textbf{Para.} & \textbf{NRMSE(\%)} & \textbf{MAE(\%)} & \textbf{R²(\%)} & \textbf{PCC(\%)} & \textbf{NRMSE(\%)} & \textbf{MAE(\%)} & \textbf{R²(\%)} & \textbf{PCC(\%)} \\
\hline

LSTM-T~\cite{pentsos2025hybrid} & 5.07M 
& 4.50±1.02 & 13.51±5.00 & 22.69±10.13 & 51.51±8.46 
& 21.20±0.24 & 29.60±0.29 & 21.70±1.70 & 47.60±0.93 \\

Informer~\cite{zhou2021informer} & 11.02M 
& 4.07±13.50 & 11.80±4.03 & 31.50±13.90 & 58.60±10.80 
& 23.87±0.47 & 30.16±1.15 & 7.47±1.90 & 39.44±0.95 \\

Crossformer~\cite{zhang2023crossformer} & 26.34M 
& 4.25±0.18 & 13.52±1.52 & 30.88±1.85 & 56.55±1.62
& 22.49±0.12 & 31.22±0.41 & 19.20±1.06 & 45.42±0.59 \\

CA-T~\cite{10.1145/3706599.3720220} & 4.54M 
& 6.55±0.42 & 8.29±1.28 & 31.91±2.12 & 63.96±4.89 
& 28.42±0.76 & 38.43±3.71 & 25.38±4.50 & 52.87±4.43 \\

PiMForce (dense)$^{\dag}$~\cite{NEURIPS2024_a01e69aa} & 66.39M
& -- & -- & -- & -- 
& 13.93±0.50 & 11.92±0.60 & 52.68±4.00 & 73.73±2.00 \\

\rowcolor{gray!10}
\textbf{NiSM (ours)} & \textbf{1.71M}
& \textbf{3.20±0.50} & \textbf{7.60±1.20} & \textbf{47.00±4.40} & \textbf{68.50±3.10}
& \textbf{11.10±0.20} & \textbf{8.69±0.30} & \textbf{54.40±2.90} & \textbf{76.60±1.40} \\
\hline
\end{tabular}
\caption{Cross-user Performance Comparison of NiSM against Baselines on Self-collected and PiMForce Datasets. (lower values are better for NRMSE/MAE, higher values are better for R²/PCC). PiMForce model $^{\dag}$ was not evaluated on our self-collected dataset because it requires dense inputs (e.g., 8-channel sEMG) that our sparse setup does not provide; instead, we report PiMForce (dense) on the public dataset using its original 8-channel EMG and treat it as a strong dense EMG reference. All other sequence baselines use the 1-ch EMG view.}
\label{tab:performance_benchmark}
\end{table*}

\subsection{Does Structure Improve Data Efficiency under Limited User Data?}

% To evaluate NiSM's generalization capability and data efficiency, we systematically varied the number of users in the training set while assessing R² performance on a fixed, unseen test set of users. As shown in Figure \ref{fig:r2_pcc_ablation}(a,b), while overall performance decreases with less training data, NiSM consistently outperforms the large dense-sensing PiMForce reference across all user counts. Under extreme data scarcity (2–5 users), NiSM retains higher correlation (PCC: 0.652 vs. 0.600) and competitive R² (0.331 vs. 0.325), despite using far fewer parameters. This robustness indicates that task-dependent modulation, sparse gating, and EMG-conditioned scaling act as effective structural regularizers, helping NiSM learn generalizable mappings from limited user data.

\begin{figure}[htbp]
  \centering
  \subfigure[R\textsuperscript{2} Trend]{
    \includegraphics[width=0.35\textwidth]{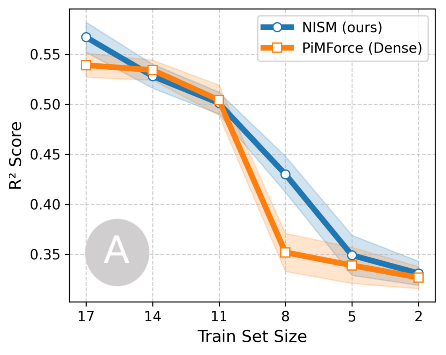}
  }\hfill
  \subfigure[PCC Trend]{
    \includegraphics[width=0.35\textwidth]{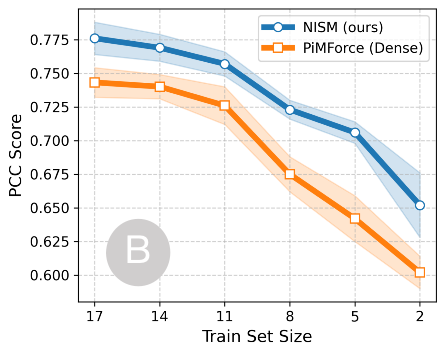}
  }\hfill
  % \subfigure[Ablation Study Results]{
  %   \includegraphics[width=0.52\textwidth]{figures/ablation.png}
  % }
  \vspace{-5pt}
  \caption{(A) R\textsuperscript{2} and (B) PCC performance trends across varying train set sizes. Shaded bands denote the standard deviation across held-out test users.}
  \label{fig:r2_pcc_ablation}
\vspace{-5pt}
\end{figure}

Sparse wearable force estimation is especially vulnerable to limited user coverage because both motion-force mappings and EMG-force relationships vary across individuals. 
Figure\ref{fig:r2_pcc_ablation}(A, B) shows that NiSM degrades more gracefully as the number of training users decreases, suggesting that the proposed structural priors act as regularizers rather than merely increasing model capacity. These results reflect zero-shot generalization to unseen users; on PiMForce, calibrating with 5\% of a test user's trials yields R\textsuperscript{2} = 64.69\% (Appendix~\ref{app:fewshot}).

\label{fig:3in1}

\subsection{Which Structural Priors Recover Missing Force-relevant Information?}

\begin{table}[htbp]
\centering
\footnotesize
\setlength{\tabcolsep}{4pt}
\renewcommand{\arraystretch}{1.05}
\begin{tabular}{lcccc}
\toprule
\textbf{Models} & \textbf{NRMSE (\%)} & \textbf{MAE (\%)} & \textbf{R$^2$} & \textbf{PCC} \\
\midrule
NiSM (full) & 10.99 & 8.50 & 0.570 & 0.773 \\
w/o Context Mod. & 11.62 & 10.06 & 0.520 & 0.754 \\
w/o SG Feedforward & 11.35 & 8.81 & 0.541 & 0.771 \\
w/o EMG Scaling & 12.08 & 9.17 & 0.464 & 0.653 \\
\bottomrule
\end{tabular}
\vspace{5pt}
\caption{Ablation study (cross-user). NRMSE and MAE are normalized and shown in percentage format.}
\label{tab:ablation}
\end{table}

% To quantify the contribution of each structural inductive bias, we performed systematic ablations on PiMForce (Figure \ref{fig:r2_pcc_ablation}(c)). Removing individual components—Intent-Conditional Modulation (ICM via CBN), Sparse Learned Gating (SLG via SGF), and EMG-conditioned Scaling—each led to clear performance degradation, confirming that these modules provide complementary constraints for force prediction under sparse sensing. Specifically, ablating ICM (CBN) lowered R², indicating that task intent is important for contextualizing sparse thumb kinematics and disambiguating how similar motions correspond to different grasp conditions. Removing SLG (SGF), which imposes sparse feature-to-finger mappings, also reduced R² (e.g., to 0.541), highlighting the value of structured kinematic coupling between thumb motion and finger-specific pressure outputs. Disabling EMG Scaling yielded the largest single-module drop, emphasizing that muscle-effort information provides an essential modulation signal for adapting base predictions to user-specific force patterns. Together, these results show that NiSM's gains do not arise merely from model capacity, but from explicitly organizing task, motion, and effort cues into structural priors. This supports the central trade-off studied in this work: under reduced sensing density, carefully designed inductive biases can recover part of the force-relevant information typically supplied by denser sensing.
We ablate NiSM's main structural components under the same sparse IMU-EMG test-time inputs. 
As shown in Table~\ref{tab:ablation}, removing context modulation lowers \(R^2\) from 0.570 to 0.520, indicating that the IMU-derived grasp context helps disambiguate similar thumb motions across hand-shape and contact states. 
Removing sparse gating also degrades performance (\(R^2=0.541\)), supporting its role in modeling finger-specific kinematic coupling. 
The largest drop occurs without EMG-conditioned scaling (\(R^2=0.464\), PCC \(=0.653\)), showing that muscle-effort modulation provides force-relevant information beyond IMU kinematics.

Together, these results show that NiSM's gains come from complementary structural priors: inferred grasp context approximates unobserved task state, sparse gating encodes kinematic coupling, and EMG scaling supplies effort-dependent information. Appendix Fig.~\ref{fig:w_sparse_visualization} shows stable feature-to-finger patterns in \(W_{\mathrm{sparse}}\), supporting that sparse gating learns structured dependencies.

Because the task condition is derived from the same thumb IMU as the kinematic feature, we further test whether it adds information beyond self-conditioning. When the task condition is swapped across actions with the other inputs fixed, learned target conditions redirect relative force allocation by 90.4\%, compared with 8.5\% for random vectors. On action pairs with matched thumb trajectories but different force patterns, task conditioning reduces force-difference recovery error by 14.74\% (Appendix~\ref{app:context}).

\section{Discussion and Limitations} 
\label{sec:discussion_limitation}

Our results support the central trade-off studied in this work: structural inductive biases can partially compensate for reduced sensing density in finger-level force estimation. With a thumb-mounted IMU and a single-channel wrist EMG sensor, NiSM is competitive with dense-sensing references while using substantially fewer parameters, showing that explicitly encoding task context, kinematic coupling, and muscle-effort modulation recovers part of the force-relevant information typically provided by denser sensing.

Several limitations remain. First, context modulation depends on IMU-to-hand-shape inference: similar hand shapes may correspond to different object properties or contact conditions, especially for unseen objects, and tasks with weak thumb--finger coupling (e.g., typing or piano playing) lie outside NiSM's scope and may require additional sensors or explicit task cues. Second, NiSM estimates per-finger normal pressure rather than 3D forces or torques, which would require vector-valued supervision such as 3D tactile or six-axis force/torque sensing. Third, participants were healthy right-handed adults with a single self-donned placement per session; in a simulated ring rotation (Appendix~\ref{app:rotation}), R\textsuperscript{2} drops by 2.6 points for random per-trial offsets up to 15$^\circ$ and by 23.4 points for a fixed 30$^\circ$ offset, so large mounting rotations call for orientation calibration; physical re-donning, wearing depth, calibration drift, and motion artifacts were not evaluated. Future work includes more expressive conditioning (e.g., cross-attention or mixture-of-experts), physically grounded priors such as joint, tendon, and contact constraints, and longer-term deployment in less constrained settings.

\begin{ack}
This work was supported in part by the National Natural Science Foundation of China (62302168, W2532047), the Guangdong S\&T Programme (2025B0101120004), the Guangdong Provincial Department of Science and Technology (2023CX10X070), the Guangdong Basic and Applied Basic Research Foundation (2026A1515010254), the Guangdong Provincial Key Laboratory of Human Digital Twin (2022B1212010004), the Guangzhou Basic and Applied Basic Research Foundation (2024A04J6311), and the Shenzhen Holdfound Foundation Endowed Professorship. This work is partially supported by the High Performance Computing Platform of South China University of Technology.

The authors declare no competing interests.
\end{ack}

\bibliographystyle{abbrv}
\bibliography{sample-base}

\newpage

\appendix

\section{Technical Appendices and Supplementary Material}
% Technical appendices with additional results, figures, graphs and proofs may be submitted with the paper submission before the full submission deadline (see above), or as a separate PDF in the ZIP file below before the supplementary material deadline. There is no page limit for the technical appendices.

\subsection{Dataset details}

\subsubsection{Self-collected Dataset}

\paragraph{Participant Recruitment}

We recruited 20 healthy right-handed adult volunteers to participate in the data collection study. All participants provided written informed consent prior to enrollment. Demographic information was anonymized in compliance with privacy standards.

Inclusion criteria included the absence of neuromuscular or musculoskeletal disorders affecting the upper limbs, and the ability to complete the full range of gesture tasks. Exclusion criteria included recent wrist or hand injuries, or the use of assistive electronic devices.

\paragraph{Ethical Approval}

All participants provided written informed consent prior to enrollment, and all procedures were conducted in accordance with approved IRB protocols.

\paragraph{Experimental Protocol}
Participants were instructed to perform 20 distinct hand gestures and grasp actions, categorized into three groups: baseline postures, rhythmic finger motion, and object manipulation.

\begin{figure}[!tbp] \centering \includegraphics[width=\textwidth]{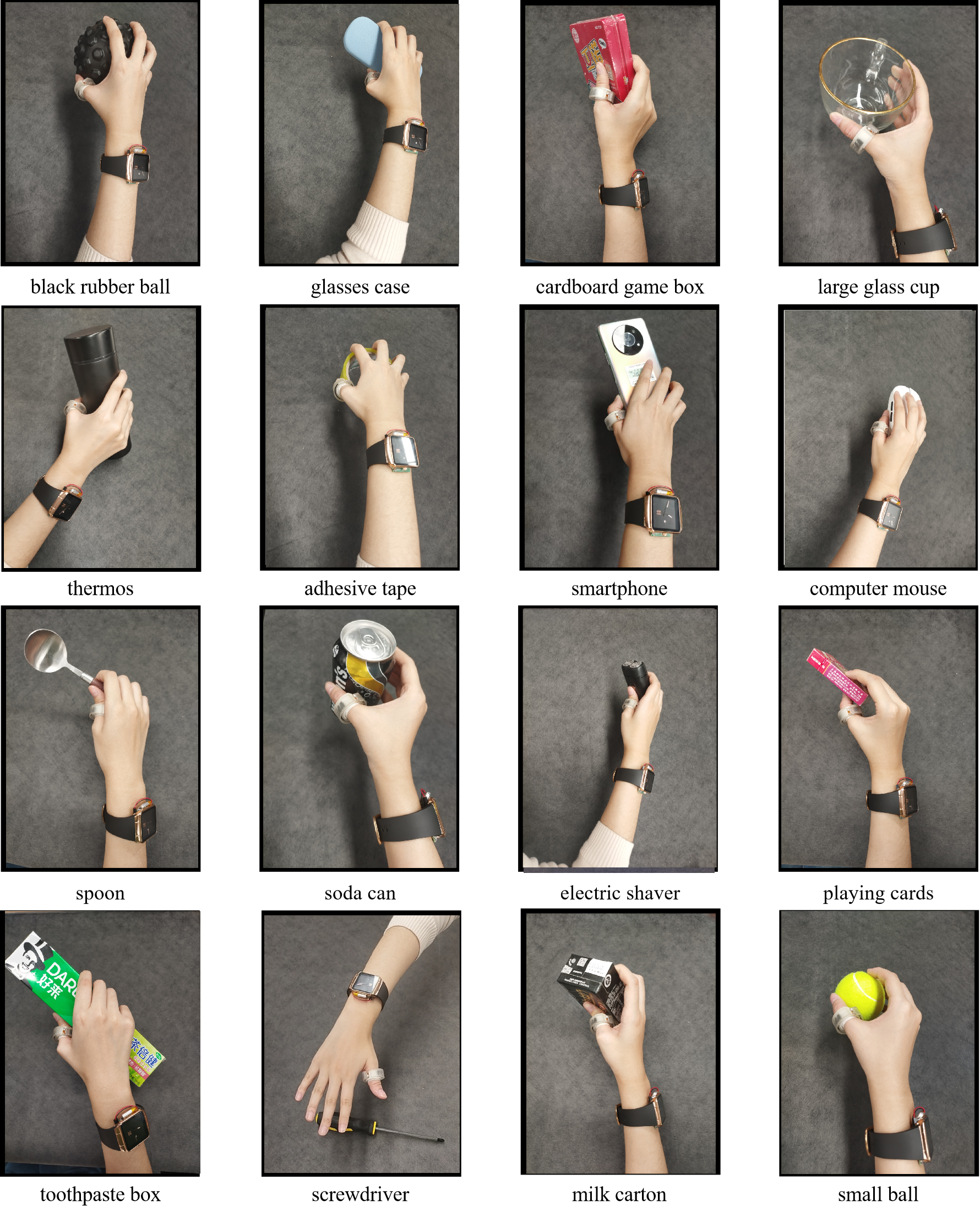}
\caption{The 16 real-world objects used in the study, selected to represent diverse physical characteristics.} \label{fig:objects}
\vspace{-10pt}
\end{figure}

\begin{itemize}
    \item \textbf{Basic Hand Gesture Tasks}
    \begin{itemize}
        \item \textbf{Palm-up rest:} Hand placed palm-up on a table with fingers naturally extended for 30 seconds.
        \item \textbf{Palm-down rest:} Hand placed palm-down with fingers loosely resting on the surface for 30 seconds.
        \item \textbf{Suspended posture:} Hand held above the table, palm facing downward, with fingers slightly flexed and avoiding contact for 30 seconds.
        \item \textbf{Rhythmic Finger Tapping:} Participants performed a rhythmic tapping task using the index, middle, and ring fingers, following a 1--2--3 pattern at a frequency of 2~Hz. Each cycle consisted of 6 seconds of tapping followed by 2 seconds of rest, repeated for 10 cycles (approximately 1 minute). The wrist was elevated to ensure motion originated exclusively from the fingers.

    \end{itemize}

    \item \textbf{Object Interaction Tasks}  
    Participants interacted with 16 real-world objects, each selected to represent diverse physical characteristics, as shown in Figure \ref{fig:objects}. For each object, participants performed five grasp-and-release repetitions, lifting the object from a table and placing it back naturally. Objects were grouped by functional context:
    \begin{itemize}
        \item \textbf{Containers and packages:} Soda can, milk carton, thermos, toothpaste box, playing cards, adhesive tape
        \item \textbf{Tools and electronics:} Screwdriver, electric shaver, smartphone, computer mouse
        \item \textbf{Personal and soft items:} Black rubber ball, small ball, glasses case
        \item \textbf{Utensils:} Large glass cup, spoon
        \item \textbf{Other:} Cardboard game box
    \end{itemize}
\end{itemize}

\paragraph{Sensor Setup}
We used a sparse wearable sensing setup for model inputs and separate fingertip sensors for ground-truth supervision. The model inputs consist of a thumb-mounted IMU and a single-channel wrist EMG sensor, while five fingertip pressure sensors provide synchronized supervision labels.

\begin{itemize}
    \item \textbf{Thumb-mounted IMU.} The system incorporates a custom-designed ring that is comfortably worn on the thumb, housing an InvenSense ICM-20948 9-axis IMU (Inertial Measurement Unit). This device integrates both an accelerometer and a gyroscope, capturing detailed movement and orientation data across three axes. The IMU samples data at a rate of 55 Hz, providing a high level of precision for motion tracking. Data is wirelessly transmitted via Bluetooth Low Energy (BLE), a protocol selected for its low power consumption, ensuring efficient long-duration use without significant battery drain. This setup captures the dynamic motion of the thumb, a key component in hand interaction, given its frequent involvement in most hand gestures. No instructions were given on ring orientation, rotational alignment, or wearing depth, and each session used a single donning.

    \item \textbf{Wrist-worn EMG sensor.} A single-channel dry-electrode EMG sensor is embedded in a smartwatch-like form factor and positioned on the dorsal forearm, near the wrist. The EMG signals, which measure electrical muscle activity, are amplified and band-pass filtered within 20--450 Hz, with notch filtering at 50 Hz to remove powerline interference. The signal is digitized at 1 kHz and transmitted via BLE to a nearby host. This sensor provides a coarse muscle-effort cue related to grasp force, complementing the thumb IMU without requiring multi-channel EMG arrays across the forearm or hand.

    \item \textbf{Fingertip pressure sensors.}
    Tekscan FlexiForce A402 sensors were placed on all five fingertips using a lightweight, flexible glove. These sensors recorded fingertip normal-force readings during interaction, which were used as ground-truth contact-pressure labels after normalization. The sensors sampled at 100 Hz and were integrated with an Arduino logger for data acquisition. These pressure labels were synchronized with the thumb IMU and wrist EMG streams using shared timestamps, alongside a host-side heartbeat signal at 100 ms intervals to maintain temporal alignment across all modalities.

\end{itemize}

\paragraph{Data Acquisition and Preprocessing}

\begin{itemize}
\item \textbf{Logging and Synchronization.} IMU and EMG signals were streamed over Bluetooth Low Energy (BLE) to a host machine, while force data were acquired via USB serial from an Arduino logger. All data streams were timestamped based on host-side receipt time. The datasets were pre-aligned before processing, assuming synchronized start and consistent sampling rates across modalities.
    
\item \textbf{Sliding Window Segmentation.}
Multimodal sensor data were segmented into sliding windows of 30 frames with a stride of 2 samples, corresponding to approximately 250 ms per segment with high temporal overlap. For each segment, the final-frame pressure vector was used as the main regression target, while pressure values within the window were retained for computing the peak-pressure loss term. The window thus contains no frames after the target, and evaluation is window-based rather than zero-latency streaming. The length $T=30$ was chosen empirically; on PiMForce, shorter windows ($T=10, 20$) reduce latency but degrade accuracy (Table~\ref{tab:window}).

\item \textbf{EMG Feature Extraction.} Each EMG segment was transformed into a fixed-length 10-dimensional feature vector using a standardized pipeline. This included common time-domain descriptors such as mean absolute value, root mean square (RMS), zero-crossing rate, waveform length, variance, amplitude sum, and standard deviation. In addition, frequency-domain features were computed via fast Fourier transform (FFT), including mean power spectral density (PSD), peak-frequency index, and total spectral energy. These features capture both transient and periodic components of muscle activity relevant to grasp and movement intention.
    
\item \textbf{Normalization and Scaling.} To reduce inter-subject variability and enable generalization, each modality was independently normalized using scalers fit on the training set. IMU and EMG features were standardized using StandardScaler, while contact-pressure labels were scaled to $[0,1]$ using MinMaxScaler per finger. The same transformations were applied during evaluation to ensure consistency.
    
\item \textbf{Final Feature Composition.} Each processed sample included four components:
    \begin{itemize}
        \item A 30-frame IMU sequence (with 12 dimensions per frame, including 3D acceleration and 9D orientation),
        \item A 10-dimensional vector of EMG features,
        \item A 5-dimensional normalized contact-pressure vector representing fingertip pressure labels.
    \end{itemize}
Optionally, the raw EMG value at the final timestep was retained for visualization or auxiliary analysis. These composite inputs formed the basis for multimodal supervised learning in our model pipeline.
\end{itemize}

\paragraph{Code and dataset} Code is available at \url{https://github.com/3525starmoon-glitch/NiSM} under the MIT License. The dataset is available upon request for non-commercial research (CC BY-NC 4.0 terms); participants are identified only by anonymous IDs, raw video is not shared, and requesters must not attempt to re-identify participants.

\subsubsection{Public Benchmark Dataset (PiMForce)}
\label{app:pimforce}

We evaluate our model on the publicly available PiMForce dataset, which contains synchronized multimodal recordings of 21 right-handed participants performing 22 grasp and manipulation tasks (indexed 0--21 in the released data). Each subject completed three sessions of repeated trials, with data stored hierarchically as \texttt{User\{ID\}/Session\{ID\}/data\_\{task\_id\}.npy}. Each file includes multimodal signals such as multi-channel EMG, inertial measurements, 3D hand pose, and fingertip/palm pressure measurements. In our evaluation, pressure measurements are used as supervision and are projected to a unified five-finger contact-pressure space.

To evaluate sparse-input models under the same sensing constraints as our ring-watch setup, we construct sparse inputs from a 12-dimensional IMU representation and a single-channel EMG signal. Specifically, the 3D acceleration and rotation matrix are concatenated to form the IMU representation, and one EMG channel is selected to match the single-channel wrist EMG setting. For the dense PiMForce reference, we use the dense modalities available in PiMForce, including multi-channel EMG and 3D hand posture, while pressure measurements remain the supervision target. All modalities are temporally aligned, normalized using training-set statistics, and segmented into overlapping windows of 30 frames.

A total of 4 users are held out for testing, while the remaining 17 users are used for training. For training users, only session 1 is included. For each test user, all three sessions are used. Out of consideration for training time, only half of the training data is used. This results in approximately 522{,}292 (half) training samples and 122{,}893 test samples. All processed data are stored as windowed numpy arrays and used consistently across model variants to ensure fair and reproducible evaluation.

\subsection{Model Implementation Details}

All NiSM experiments except the rotation simulation (Appendix~\ref{app:rotation}), which used one NVIDIA A800 GPU, were conducted on a system equipped with an Intel(R) Xeon(R) Gold 6138 CPU @ 2.00GHz and an NVIDIA GeForce RTX 4090 GPU; baselines whose hardware differs are noted in their own paragraphs below. The deep learning framework utilized was PyTorch, version 2.5.1, running in a Python  3.9 environment.

\subsubsection{NiSM Architecture Details}

The neuro-inspired structured architecture projects sparse multimodal inputs into a shared 48-dimensional latent space (\(d_z = 48\)). 
The IMUEncoder processes 12-dimensional IMU sequences using a two-layer bidirectional LSTM and outputs a 48-dimensional kinematic feature. 
In parallel, a HandShapeLatentEncoder maps the IMU representation to an IMU-derived hand-shape latent, which is further transformed by a two-layer MLP-based ContextEncoder into a 48-dimensional inferred grasp-context feature. 
When object/action labels are available during training, an auxiliary classification head is attached to this context feature, but these labels are not embedded or provided to the force predictor at inference.

The structured prediction modules then use this inferred context feature to condition force estimation. 
Conditional Batch Normalization normalizes the IMU feature, with scale \(\gamma\) and shift \(\beta\) dynamically predicted from the grasp-context feature. 
Base pressure predictions for five fingers are generated through a sparse feature-to-finger projection regularized by the \(\ell_1\) sparsity term described in Section~\ref{sec:training_loss}. 
The predictions are further gated by a ContextMask module conditioned on the inferred grasp-context feature. 
Finally, the EMGScalingModule maps the 48-dimensional EMG feature to five finger-wise scaling factors using a two-layer MLP with ReLU activation and 0.2 dropout, followed by a Sigmoid activation to produce \(\mathbf{s}_{\mathrm{EMG}} \in [0,1]^5\).

\subsubsection{Baseline Architecture Details}
\paragraph{LSTM-Trans}
\begin{itemize}
    \item \textbf{Model Implementation.}We adopt a hybrid Transformer-LSTM model as a baseline to infer multi-channel finger pressure signals from multimodal sensor inputs, including electromyography (EMG) and inertial measurement unit (IMU) signals. The model architecture and training configuration follow the original implementation without modification and serve as a reference point for evaluating our proposed method. Each input sample is a fixed-length sequence of 30 time steps, where each time step consists of concatenated EMG and IMU features. The model first processes the input sequence using two stacked LSTM layers (each with 128 hidden units), which extract temporal dependencies from the raw multimodal signals. The LSTM output is passed through two Transformer encoder blocks, each comprising a multi-head self-attention mechanism (4 heads, 64-dimensional each), dropout (rate = 0.2), and a position-wise feed-forward convolutional module (channels: 128 and 64) with ReLU activation. Layer normalization and residual connections are used to enhance stability and convergence. The output is then passed through a series of fully connected layers and a final linear projection layer to regress the target finger pressure sequence across 5 pressure channels.
    \item \textbf{Training Configuration.}The model is trained using the Adam optimizer with an initial learning rate of $1 \times 10^{-3}$. The loss function is mean squared error (MSE), and evaluation metrics include root mean squared error (RMSE), mean absolute error (MAE), mean absolute percentage error (MAPE), coefficient of determination ($R^2$), Pearson correlation coefficient (PCC), and Spearman correlation. We use a batch size of 64 and train for up to 50 epochs with early stopping and learning rate scheduling based on validation loss. Input and output signals are min-max normalized before training, and the model’s predictions are inverse-transformed for evaluation.This model is used solely as a baseline under the original parameter setting, without tuning or structural modifications.
\end{itemize}

\paragraph{Informer}
\begin{itemize}
    \item \textbf{Model Implementation.}We build upon the official Informer architecture, originally designed for long-term time-series forecasting. The source code implements an encoder-decoder framework with ProbSparse attention for $\mathcal{O}(L \log L)$ time complexity. We retain the original modular design, including the \textit{Encoder}, \textit{Decoder}, \textit{DataEmbedding}, \textit{AttentionLayer}, and \textit{ConvLayer}, while only making minimal task-specific modifications to accommodate our finger-pressure inference task.
    
    Specifically, we adapt the model to the short-term forecast setting \textit{$(task name='short term forecast')$} by using synchronized IMU and EMG sequences as input and finger pressure as target. Each time step of input uses the same sparse IMU and EMG view as the other baselines, and the model is trained to predict a 5-dimensional finger pressure vector for each output time step. The encoder and decoder are initialized with two layers each, 4 attention heads, 128-dimensional model embeddings, and 512 feed-forward dimensions. The sinusoidal positional embedding and convolutional distillation layers are preserved. Input and target sequences are scaled using per-sample z-score normalization before encoding, and inverse transformed after decoding.The original forward pass structure is preserved. The \textit{short forecast} method is used as the primary inference function. The \textit{long forecast}, imputation, and classification branches are disabled in our task. No attention visualization or output-attention flag is used in final evaluation.

    \item \textbf{Training Configuration.}The model is trained using the Adam optimizer with an initial learning rate of $5 \times 10^{-4}$, decayed by a factor of 0.1 upon validation loss stagnation. The training loss is mean squared error (MSE), and the evaluation metrics include RMSE, MAE, MAPE, coefficient of determination ($R^2$), Pearson correlation coefficient (PCC), and Spearman correlation coefficient. A batch size of 32 is used, and training is conducted for up to 80 epochs with early stopping (patience = 10) and checkpointing based on validation RMSE. Both input sensor data and ground truth finger pressure values are normalized to [0, 1] using min-max scaling before training, and inverse transformations are applied to predictions for evaluation. All experiments are conducted on a single NVIDIA RTX 3090 GPU. No parameter tuning or additional modules are introduced beyond the minimal modifications required for the task.
\end{itemize}

\paragraph{Crossformer}
\begin{itemize}
    \item \textbf{Model Implementation.}We build upon the official Crossformer architecture~\cite{zhang2023crossformer}, which segments each input variable into patches and applies two-stage attention across the time and variable dimensions to capture cross-dimension dependencies. We retain the original modular design and only make minimal task-specific modifications so that the model takes the same synchronized EMG and IMU windows (30 time steps) as the other baselines and outputs the 5-dimensional finger pressure.
    \item \textbf{Training Configuration.}The model is trained with the default hyperparameters of the official implementation, without additional tuning, and is evaluated on the same data splits and metrics as the other baselines.
\end{itemize}

\paragraph{PiMForce-Dense.}
\begin{itemize}
    \item \textbf{Model Implementation.} As a dense-sensing reference, we retain most of the original PiMForce architecture with minimal adaptations to our five-finger pressure evaluation. The model estimates finger contact pressure from the dense sensing modalities available in PiMForce, including multi-channel EMG and 3D hand posture, while pressure measurements are used only as supervision. The model adopts a dual-stream structure combining a ResNet-50 backbone for pose encoding and a customized EMGNet module for EMG signal processing.

    In parallel, 3D hand-pose sequences are processed using a ResNet-50 network to extract a 512-dimensional pose representation. The two embeddings are concatenated to form a 1024-dimensional joint feature, which is passed through a three-layer MLP with hidden dimensions of 256 and 256, and finally projected to a vector of size $F_L \times F \times 32$, where $F_L$ is the number of force levels and $F$ is the number of finger force channels. All layers in the MLP use ReLU activations and are followed by Batch Normalization. The final output is reshaped to $(B, F_L, F, 32)$, where $B$ denotes the batch size, providing a structured representation for downstream pressure regression tasks.

    \item \textbf{Training Configuration.} Starting from the published hyperparameters, we searched the learning rate and batch size on a validation split of the training data; the values below are the selected ones. The Adam optimizer is employed with an initial learning rate of $1 \times 10^{-4}$, and the learning rate is reduced by a factor of 0.2 if the monitored metric plateaus (the training loss in the fixed split and the validation RMSE in the leave-one-subject-out evaluation). Mean Absolute Error (MAE) is chosen as the loss function. The model is evaluated using metrics such as Root Mean Squared Error (RMSE), Mean Absolute Percentage Error (MAPE), Pearson Correlation Coefficient (PCC), Spearman Rank Correlation, and the coefficient of determination $R^2$. The training process uses a batch size of 64 for up to 100 epochs. In the fixed split, the final-epoch model is used; in the leave-one-subject-out evaluation, training stops after 15 epochs without improvement, and the checkpoint with the lowest RMSE on the separate validation participant is used. Both input sensor data and the ground truth are normalized to the $[0, 1]$ range using min-max scaling prior to training, and inverse scaling is applied to predictions during evaluation. All experiments are performed on a single NVIDIA A800 GPU with 80GB of VRAM. No architectural adjustments are made beyond the minimal task-specific changes.
\end{itemize}

\paragraph{CA-T}
\begin{itemize}
    \item \textbf{Model Implementation.} The model is a multimodal Transformer-based architecture designed for joint prediction of fingertip force and hand pose from wearable IMU and EMG inputs. The IMU stream is processed using a 1D convolutional encoder followed by a 2-layer Transformer encoder with 4 attention heads and 128-dimensional hidden states. In parallel, the EMG signal is encoded using a 2-layer LSTM with the same hidden size. To enable cross-modal reasoning, we insert bidirectional cross-attention modules between the IMU and EMG branches. The resulting fused representations are decoded through two Transformer decoders—one for 3D hand pose and another for fingertip pressure—each comprising 2 layers and 4 heads. Final predictions are obtained through task-specific MLP heads. In addition to the main outputs, the model includes auxiliary branches for unimodal predictions (IMU-only, EMG-only), and a cross-attention-based predictor, enabling multi-level supervision across all intermediate stages.

    \item \textbf{Training Configuration.} The model is trained using the Adam optimizer with an initial learning rate of $1 \times 10^{-3}$, which is decayed by a factor of 0.5 every 30 epochs. During fine-tuning, a smaller learning rate of $1 \times 10^{-4}$ and longer decay schedule (every 100 epochs) are applied. We use a batch size of 256 and train for up to 100 epochs. Training is supervised by a composite objective comprising MSE loss for pose prediction and Smooth L1 loss for pressure estimation. Auxiliary predictions from the unimodal branches and cross-attention module are also supervised with weighted losses. The loss weights are empirically set as: 1.0 for final pose, 3.0 for final pressure, 0.3 for unimodal pose heads, 0.6 for unimodal pressure heads, and 0.5 for the cross-attention output. All models are trained and evaluated on a single NVIDIA GPU using PyTorch. In the fixed split, the final-epoch model is used; in the leave-one-subject-out evaluation, early stopping (patience 50 epochs) is applied based on the loss of the separate validation participant.
\end{itemize}

\subsection{Training Procedure}
The model was trained utilizing the Adam optimizer, configured with a learning rate of $1 \times 10^{-4}$ and a weight decay of $1 \times 10^{-5}$; the Adam optimizer's beta parameters defaulted to standard PyTorch values. A ReduceLROnPlateau learning rate scheduling strategy was employed, which monitored the training loss in the fixed split, where no held-out data were used, and the loss of the separate validation participant in the leave-one-subject-out evaluation. The learning rate was reduced by a factor of $0.5$ if the monitored metric failed to demonstrate an improvement of at least $1 \times 10^{-3}$ over 3 consecutive epochs. Training was conducted with a batch size of $256$ for $100$ epochs, and the final-epoch model was used; in the leave-one-subject-out evaluation (Section~\ref{sec:dense_gap}), the checkpoint was instead selected on a separate non-test participant.

The composite loss function incorporated a Mean Squared Error (MSE) task loss, $L_1$ sparsity regularization terms for the SparseGatedFeedforward module ($\lambda_{SGF_{sparse}} = 0.0001$) and the context-mask module ($\lambda_{\mathrm{mask}_{sparse}}$ $= 0.0001$), alongside a peak-weighted error component. This peak-weighted error component, designed to penalize under-predictions at signal peaks—defined as instances where ground truth pressure exceeded $0.7$—was assigned a weight of $\lambda_{peak} = 2$.

\subsection{Evaluation Details}

\subsubsection{Evaluation Metrics}

We report four standard regression metrics to quantitatively evaluate the accuracy of finger force prediction: Normalized Root Mean Square Error (\textbf{NRMSE}), Mean Absolute Error (\textbf{MAE}), Coefficient of Determination ($\boldsymbol{R}^2$), and Pearson Correlation Coefficient (\textbf{PCC}). These metrics are computed on a per-frame, per-finger basis and averaged over the entire test set.

Let $\hat{\mathbf{y}}_i \in \mathbb{R}^d$ denote the predicted force vector for sample $i$, and $\mathbf{y}_i \in \mathbb{R}^d$ the corresponding ground truth, where $d = 5$ is the number of fingers. Let $N$ be the number of test samples.

\textbf{MAE} measures the average magnitude of absolute errors, regardless of direction:
\begin{equation}
\text{MAE} = \frac{1}{N} \sum_{i=1}^{N} \frac{1}{d} \sum_{j=1}^{d} \left| \hat{y}_{ij} - y_{ij} \right|
\end{equation}

\textbf{NRMSE} provides a scale-invariant measure of prediction error, normalized by the dynamic range of the ground truth per finger:
\begin{equation}
\text{NRMSE} = \frac{1}{d} \sum_{j=1}^{d} \frac{\sqrt{\frac{1}{N} \sum_{i=1}^{N} \left( \hat{y}_{ij} - y_{ij} \right)^2}}{\max(y_{\cdot j}) - \min(y_{\cdot j})}
\end{equation}

$\boldsymbol{R}^2$ quantifies the proportion of variance in the ground truth that is explained by the model predictions:
\begin{equation}
R^2 = 1 - \frac{\sum_{i=1}^{N} \sum_{j=1}^{d} \left( \hat{y}_{ij} - y_{ij} \right)^2}{\sum_{i=1}^{N} \sum_{j=1}^{d} \left( y_{ij} - \bar{y}_j \right)^2}
\end{equation}
where $\bar{y}_j = \frac{1}{N} \sum_{i=1}^{N} y_{ij}$ is the mean force value for the $j$-th finger.

\textbf{PCC} evaluates the linear correlation between predicted and true values, averaged across fingers:
\begin{equation}
\text{PCC} = \frac{1}{d} \sum_{j=1}^{d} \frac{\sum_{i=1}^{N} \left( \hat{y}_{ij} - \bar{\hat{y}}_j \right) \left( y_{ij} - \bar{y}_j \right)}{\sqrt{\sum_{i=1}^{N} \left( \hat{y}_{ij} - \bar{\hat{y}}_j \right)^2} \sqrt{\sum_{i=1}^{N} \left( y_{ij} - \bar{y}_j \right)^2}}
\end{equation}
Here, $\bar{\hat{y}}_j$ and $\bar{y}_j$ denote the mean predicted and ground-truth forces for the $j$-th finger, respectively.

\medskip

These metrics provide a comprehensive view of both absolute accuracy (MAE, NRMSE) and relative prediction quality ($R^2$, PCC). All results reported in the main paper were computed using the above formulations without modification.

\subsubsection{Paired Statistical Comparison}
\label{app:stats}
For the leave-one-subject-out comparison in Section~\ref{sec:dense_gap}, all compared models are retrained from scratch in every fold under the same partition and evaluated on the same held-out subject. Each model is trained once per fold with a fixed random seed, so the resulting statistics reflect between-subject rather than initialization variability. For R\textsuperscript{2} and PCC, let $d_i$ be the paired difference on subject $i$ ($i = 1,\dots,n$, $n = 21$), computed as NiSM minus the compared model so that positive values favor NiSM, and let $\bar{d}$ and $s_d$ be their mean and standard deviation. We report the 95\% confidence interval and the effect size
\begin{equation*}
\bar{d} \pm t_{0.975,\,n-1}\,\frac{s_d}{\sqrt{n}}, \qquad d_z = \frac{\bar{d}}{s_d},
\end{equation*}
together with the number of subjects for which NiSM is better and the two-sided Wilcoxon signed-rank $p$-value. Results are summarized in Table~\ref{tab:paired}.

\begin{table}[h]
\centering
\footnotesize
\setlength{\tabcolsep}{2.5pt}
\begin{tabular}{llcccc}
\toprule
\textbf{Comparison} & \textbf{Metric} & \textbf{Improvement [95\% CI]} & $\boldsymbol{d_z}$ & \textbf{Subjects improved} & \textbf{Wilcoxon $p$} \\
\midrule
\multirow{2}{*}{NiSM vs.\ CA-T} & R\textsuperscript{2} gain & +0.300 [0.226, 0.374] & 1.84 & 21/21 & $<0.001$ \\
 & PCC gain & +0.193 [0.157, 0.229] & 2.41 & 21/21 & $<0.001$ \\
\midrule
\multirow{2}{*}{NiSM vs.\ PiMForce (dense)} & R\textsuperscript{2} gain & +0.167 [0.128, 0.206] & 1.96 & 20/21 & $<0.001$ \\
 & PCC gain & +0.113 [0.089, 0.137] & 2.18 & 21/21 & $<0.001$ \\
\bottomrule
\end{tabular}
\vspace{5pt}
\caption{Paired comparison over 21 leave-one-subject-out folds on PiMForce. Positive values favor NiSM.}
\label{tab:paired}
\end{table}

\subsubsection{Analysis of the Inferred Task Condition}
\label{app:context}
Using the trained model without retraining, we test whether the inferred task condition has a target-specific effect and whether it helps when thumb trajectories are ambiguous.

\textbf{Task-condition swaps.} Across all $22\times21=462$ ordered non-identity action swaps on PiMForce, we fix the model, $\mathbf{z}_{\text{IMU}}$, and $\mathbf{z}_{\text{EMG}}$, and inject either the learned target condition or a task-agnostic random vector. The task-condition redirection ratio (TCR) measures source$\rightarrow$target progress of the prediction (0\% = source, 100\% = target); gap weighting accounts for source--target separation. Learned swaps strongly redirect relative force allocation, whereas random vectors have little effect; the effect is weaker but still clear for total-force magnitude (Table~\ref{tab:tcr}).

\begin{table}[h]
\centering
\footnotesize
\setlength{\tabcolsep}{3pt}
\begin{tabular}{llcccc}
\toprule
\textbf{Condition} & \textbf{Output} & \textbf{Gap-weighted TCR [95\% CI]} & \textbf{Median} & \textbf{$\geq$80\%} & \textbf{$\geq$95\%} \\
\midrule
Learned target & Relative force allocation & 90.4\% [88.8\%, 91.7\%] & 89.6\% & 81.2\% & 17.1\% \\
Learned target & Total-force magnitude & 56.9\% [46.7\%, 64.9\%] & 51.4\% & 21.2\% & 1.5\% \\
Random vector & Relative force allocation & 8.5\% [4.0\%, 13.2\%] & 5.8\% & 1.1\% & 0.0\% \\
Random vector & Total-force magnitude & $-$1.3\% [$-$13.3\%, 9.7\%] & $-$8.0\% & 5.6\% & 0.0\% \\
\bottomrule
\end{tabular}
\vspace{5pt}
\caption{Learned task-condition swaps versus a random-vector control on PiMForce (462 ordered action swaps). TCR: 0\% = source, 100\% = target.}
\label{tab:tcr}
\end{table}

\textbf{Matched thumb trajectories.} We select action pairs whose thumb trajectories are nearly identical (palm-normalized DTW $\leq 0.015$) but whose force patterns differ; all eligible pairs are included (3 pairs, 168 samples). Recovery error measures the mismatch between the predicted and true force difference within a pair. Task conditioning reduces recovery error from 0.81 to 0.69 ($-$14.74\%), with improvement in every pair (Table~\ref{tab:dtw}).

\begin{table}[h]
\centering
\footnotesize
\begin{tabular}{lccc}
\toprule
\textbf{Action pair} & \textbf{Samples} & \textbf{DTW} & \textbf{Recovery error ($\mathbf{z}_{\text{IMU}}$-context $\rightarrow$ task condition)} \\
\midrule
TI-Pinch vs.\ TM-Pinch & 58 & 0.01 & 0.83 $\rightarrow$ 0.71 ($-$14.51\%) \\
Sphere 3 Finger vs.\ Quadpod & 59 & 0.01 & 1.03 $\rightarrow$ 0.87 ($-$15.52\%) \\
M-Press vs.\ P-Press & 51 & 0.01 & 0.52 $\rightarrow$ 0.45 ($-$13.37\%) \\
\midrule
Overall & 168 & 0.01 & 0.81 $\rightarrow$ 0.69 ($-$14.74\%) \\
\bottomrule
\end{tabular}
\vspace{5pt}
\caption{Task conditioning improves force-difference recovery under matched thumb trajectories. DTW is palm-normalized (lower = more similar); lower recovery error is better.}
\label{tab:dtw}
\end{table}

\subsubsection{Few-shot User Calibration}
\label{app:fewshot}
Because EMG is user-dependent, we complement the zero-shot cross-user evaluation with few-shot calibration on PiMForce: a fraction of each test user's trials is used for calibration, and non-overlapping trials of the same user are used for evaluation. Results are shown in Table~\ref{tab:fewshot}, where the 0\% row is the zero-shot setting. Most metrics improve with calibration, but MAE is not strictly monotonic from 5\% to 20\%.

\begin{table}[h]
\centering
\footnotesize
\begin{tabular}{lcccc}
\toprule
\textbf{Calibration} & \textbf{NRMSE (\%)} & \textbf{MAE (\%)} & \textbf{R\textsuperscript{2} (\%)} & \textbf{PCC (\%)} \\
\midrule
0\% & 11.10 $\pm$ 0.20 & 8.69 $\pm$ 0.30 & 54.40 $\pm$ 2.90 & 76.60 $\pm$ 1.40 \\
5\% & 9.07 $\pm$ 0.08 & 6.08 $\pm$ 0.68 & 64.69 $\pm$ 7.90 & 81.76 $\pm$ 3.99 \\
20\% & 5.75 $\pm$ 0.71 & 6.84 $\pm$ 1.27 & 67.90 $\pm$ 8.46 & 83.37 $\pm$ 4.27 \\
\bottomrule
\end{tabular}
\vspace{5pt}
\caption{Few-shot calibration on PiMForce using a fraction of each test user's trials.}
\label{tab:fewshot}
\end{table}

\begin{table}[h]
\centering
\footnotesize
\begin{tabular}{lcccc}
\toprule
\textbf{$T$} & \textbf{NRMSE (\%)} & \textbf{MAE (\%)} & \textbf{R\textsuperscript{2} (\%)} & \textbf{PCC (\%)} \\
\midrule
10 & 18.84 $\pm$ 0.27 & 11.36 $\pm$ 0.58 & 38.29 $\pm$ 0.49 & 70.55 $\pm$ 0.67 \\
20 & 18.43 $\pm$ 0.28 & 10.42 $\pm$ 0.20 & 41.20 $\pm$ 0.60 & 72.83 $\pm$ 0.20 \\
30 & 11.10 $\pm$ 0.20 & 8.69 $\pm$ 0.30 & 54.40 $\pm$ 2.90 & 76.60 $\pm$ 1.40 \\
\bottomrule
\end{tabular}
\vspace{5pt}
\caption{Effect of the temporal window length $T$ on PiMForce; $T=30$ is the default setting.}
\label{tab:window}
\end{table}

\subsubsection{Simulated Ring Rotation}
\label{app:rotation}
We simulate ring rotation by rotating the IMU inputs of a trained model at test time with a mounting offset $R_{\mathrm{off}}$ ($\mathbf{a} \rightarrow R_{\mathrm{off}}^{\top}\mathbf{a}$, $R \rightarrow R\,R_{\mathrm{off}}$). The model is trained within participants on the self-collected data: each trial is split in time into 60\% training, 10\% checkpoint selection, and 30\% testing; pressure is zeroed by each participant's no-contact offset, and EMG features are averaged over 100-sample sub-windows. Without rotation, the model reaches R\textsuperscript{2} = 23.30\%. Random per-trial offsets of up to 15$^\circ$/30$^\circ$ reduce R\textsuperscript{2} by 2.6/8.5 points, and fixed offsets of 10$^\circ$/20$^\circ$/30$^\circ$/45$^\circ$ about random axes reduce it by 3.9/12.5/23.4/41.9 points (averaged over three random draws). NiSM thus tolerates small per-trial offsets, whereas large fixed offsets call for orientation calibration.

\subsubsection{Per-Finger Performance Analysis}
The provided performance metrics reveal variations in model accuracy across different fingers. Specifically, the thumb (Finger 1) achieves the lowest normalized MAE, suggesting highly accurate predictions, whereas the error gradually increases for fingers 2 through 5, peaking at Finger 5. The Pearson Correlation Coefficient (PCC) and coefficient of determination (R²) consistently remain high across all fingers, indicating robust linear relationships and good explanatory power of the model, though slightly declining from thumb to pinky. NRMSE remains stable and relatively low, further confirming the model's overall reliability in predicting finger-level forces, albeit with incremental degradation from the thumb towards the little finger.

\begin{figure}[h]
  \centering
  \includegraphics[width=\linewidth]{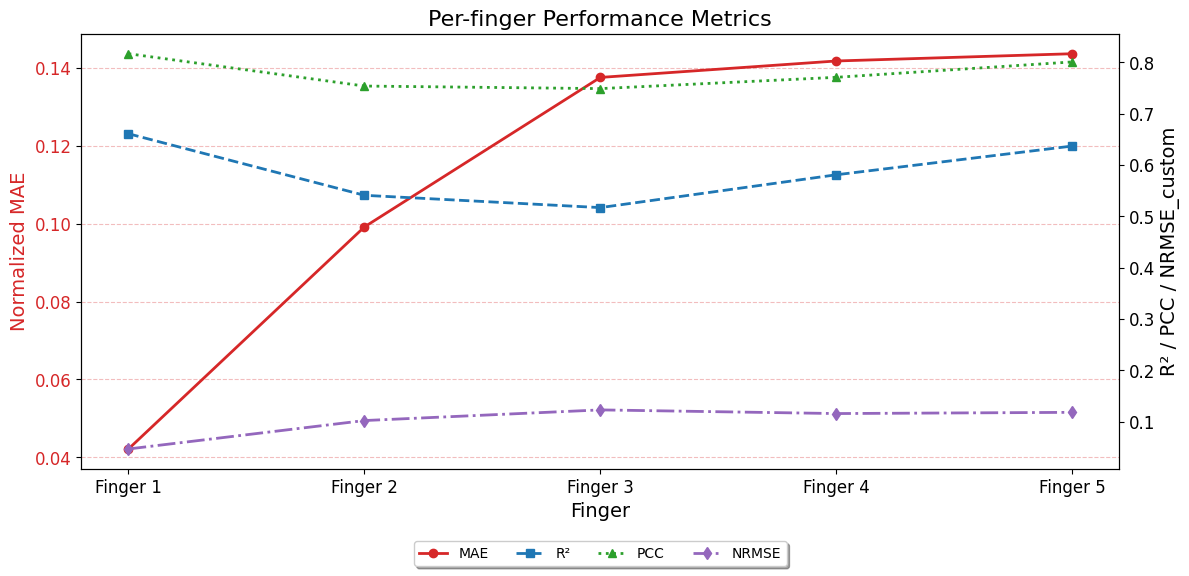}
  \caption{Per-finger performance metrics of the model for finger force prediction. The plot displays Normalized Mean Absolute Error (MAE) on the primary (left) y-axis, and $R^2$ score, Pearson Correlation Coefficient (PCC), and custom Normalized Root Mean Square Error ($NRMSE_{\text{custom}}$) on the secondary (right) y-axis, for each of the five fingers.}
  \label{fig:per_finger_metrics}
  \vspace{-10pt}
\end{figure}

\subsubsection{Qualitative Examples of Force Prediction}
\begin{figure}[h]
  \centering
  \includegraphics[width=\linewidth]{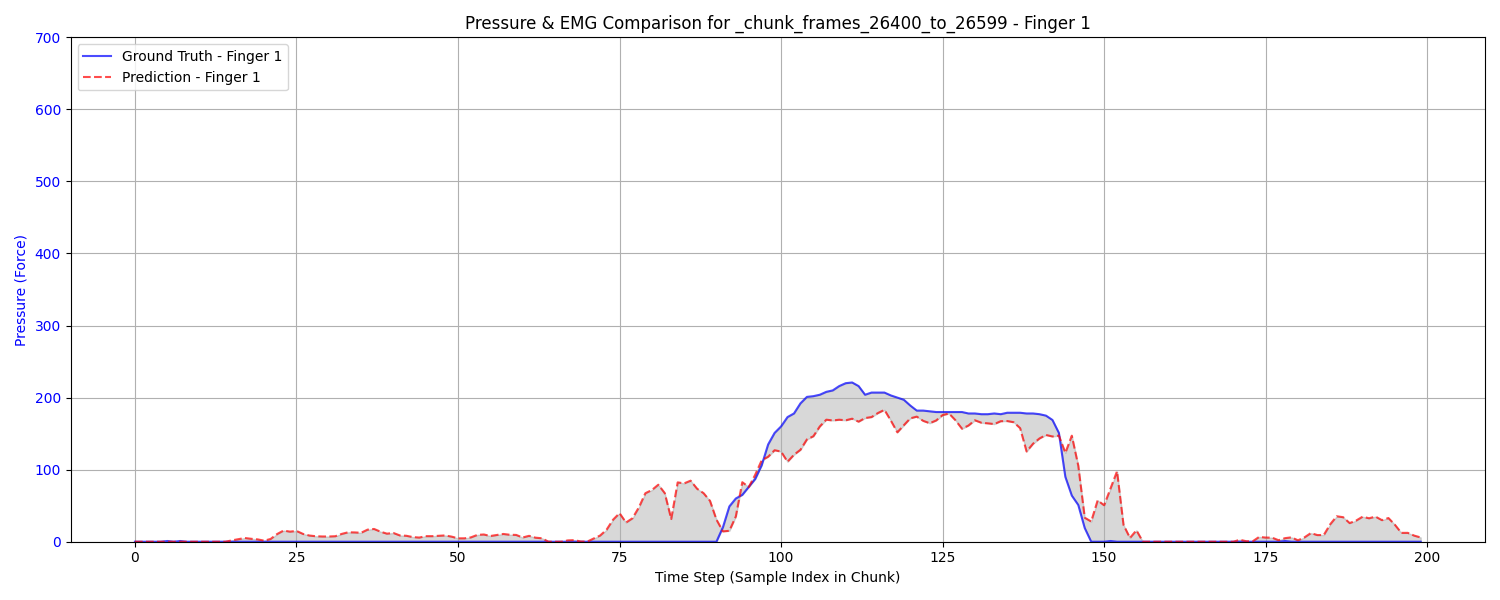}
  \caption{Example of predicted versus ground truth pressure for Finger 1 over a specific time window (frames 26400-26599). The ground truth is shown as a blue solid line, the model's prediction as a red dashed line, and the shaded area indicates the magnitude of the prediction error.}
\label{fig:pressure_comparison_f1_chunk_26400_26599}
  \vspace{-10pt}
\end{figure}

The Figure \ref{fig:pressure_comparison_f1_chunk_26400_26599} presents a qualitative comparison between predicted and ground-truth finger forces for Finger 1 across a representative time segment. The predicted force closely tracks the general trend and dynamic variations of the ground truth, effectively capturing both the onset and offset phases of applied pressure. Although minor deviations are observable at peak forces, the model successfully reflects the temporal structure of finger pressure, underscoring its effectiveness in realistic force estimation scenarios.

\subsubsection{Visualization of Learned Synergies/Weights}

\begin{figure}[h]
  \centering
  \includegraphics[width=\linewidth]{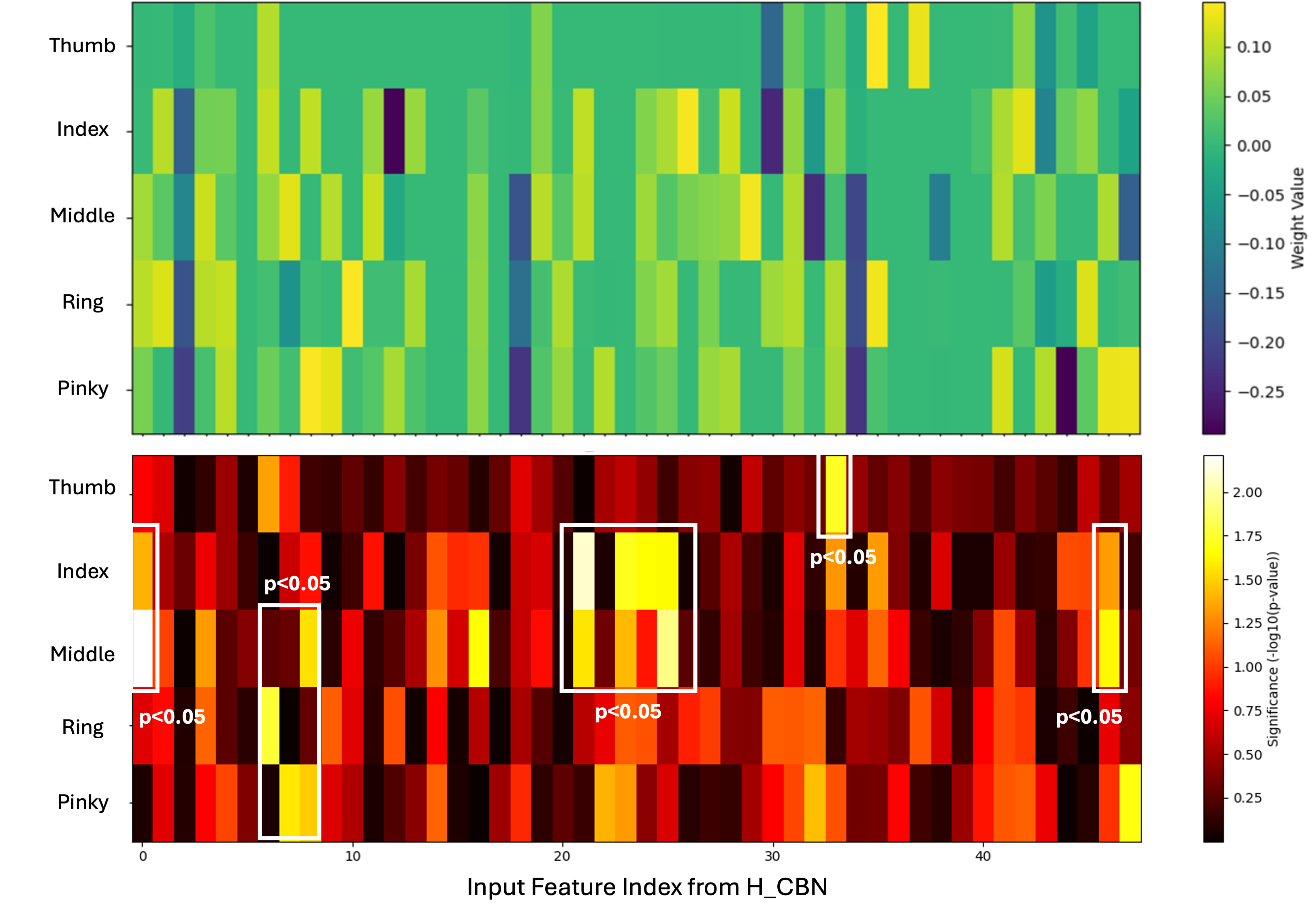}
  \caption{Analysis of the learned sparse weight matrix $W_\text{sparse}$. (Top) Mean weight values across all cross-validation folds. (Bottom) Statistical significance of each weight, visualized as -log10(p-value) from a one-sample t-test.}
  \label{fig:w_sparse_visualization}
  \vspace{-10pt}
\end{figure}

Figure~\ref{fig:w_sparse_visualization} shows the learned weight matrix $W_{\text{sparse}}$ from the Sparse Gated Feedforward (SGF) module, which maps context-modulated kinematic features ($H_{\text{CBN}}$) to per-finger base force predictions ($P_{\text{base}}$). The matrix is notably sparse—many weights are near-zero (green)—as encouraged by L1 regularization. This sparsity reveals that the model avoids dense, all-to-all mappings and instead identifies finger-specific subsets of relevant input features.

Such structure enables efficient and interpretable computation, consistent with sparse coding observed in biological motor circuits. For example, Finger 2 exhibits strong negative weights at feature indices 12–13 and strong positive weights at 29–30 and 43–44, indicating distinct, opposing influences. Meanwhile, features 32–34 negatively influence Fingers 1, 3, and 4, but have little effect on others. These patterns suggest that $H_{\text{CBN}}$ encodes higher-level kinematic or intent features that coordinate multi-finger force outputs in a synergy-like fashion.

Overall, the interpretable and disentangled structure learned by $W_{\text{sparse}}$ demonstrates the benefit of incorporating neuro-inspired structural inductive biases. It improves generalization from sparse data and offers transparency crucial for reliable human-AI interaction.

\subsection{Broader Impact Statement}
Our work advances the field of force-aware human-computer interaction by enabling accurate finger-level force estimation from sparse, wearable biosignals. This has the potential to significantly reduce the cost, intrusiveness, and power consumption of systems for dexterous input in domains such as virtual reality, assistive robotics, prosthetics, and ergonomic monitoring. By mimicking structural principles from biological motor control, our approach not only improves data efficiency and generalization but also enhances model interpretability—an important step toward trustworthy AI in embodied applications.

In the assistive technology space, our system could empower individuals with motor impairments by enabling intuitive control of prosthetic limbs or exoskeletons with minimal instrumentation. Similarly, in industrial and medical settings, the ability to monitor hand force non-invasively could help prevent repetitive strain injuries or enable remote physical therapy.

However, as with any sensing system that interfaces closely with the human body, there are potential risks related to privacy and fairness. For instance, wearable biosignal data could unintentionally reveal sensitive health conditions if not properly safeguarded. Moreover, although our model demonstrates strong generalization, it may still underperform for users whose neuromuscular characteristics are underrepresented in the training data. Addressing this will require diverse participant recruitment and careful evaluation across demographics.

Finally, our work does not support applications that attempt to infer intent or behavior covertly, or in ways that violate user consent. We strongly advocate for transparent use, informed consent, and user control in all deployments of human sensing technologies.

\newpage
\section*{NeurIPS Paper Checklist}

\begin{enumerate}

\item {\bf Claims}
    \item[] Question: Do the main claims made in the abstract and introduction accurately reflect the paper's contributions and scope?
    \item[] Answer:  \answerYes{} % Replace by \answerYes{}, \answerNo{}, or \answerNA{}.
    \item[] Justification: Section \ref{sec:introduction} includes the main claims.
    \item[] Guidelines:
    \begin{itemize}
        \item The answer \answerNA{} means that the abstract and introduction do not include the claims made in the paper.
        \item The abstract and/or introduction should clearly state the claims made, including the contributions made in the paper and important assumptions and limitations. A \answerNo{} or \answerNA{} answer to this question will not be perceived well by the reviewers. 
        \item The claims made should match theoretical and experimental results, and reflect how much the results can be expected to generalize to other settings. 
        \item It is fine to include aspirational goals as motivation as long as it is clear that these goals are not attained by the paper. 
    \end{itemize}

\item {\bf Limitations}
    \item[] Question: Does the paper discuss the limitations of the work performed by the authors?
    \item[] Answer: \answerYes{} % Replace by \answerYes{}, \answerNo{}, or \answerNA{}.
    \item[] Justification: Section \ref{sec:discussion_limitation} includes the limitations of the work.
    \item[] Guidelines:
    \begin{itemize}
        \item The answer \answerNA{} means that the paper has no limitation while the answer \answerNo{} means that the paper has limitations, but those are not discussed in the paper. 
        \item The authors are encouraged to create a separate ``Limitations'' section in their paper.
        \item The paper should point out any strong assumptions and how robust the results are to violations of these assumptions (e.g., independence assumptions, noiseless settings, model well-specification, asymptotic approximations only holding locally). The authors should reflect on how these assumptions might be violated in practice and what the implications would be.
        \item The authors should reflect on the scope of the claims made, e.g., if the approach was only tested on a few datasets or with a few runs. In general, empirical results often depend on implicit assumptions, which should be articulated.
        \item The authors should reflect on the factors that influence the performance of the approach. For example, a facial recognition algorithm may perform poorly when image resolution is low or images are taken in low lighting. Or a speech-to-text system might not be used reliably to provide closed captions for online lectures because it fails to handle technical jargon.
        \item The authors should discuss the computational efficiency of the proposed algorithms and how they scale with dataset size.
        \item If applicable, the authors should discuss possible limitations of their approach to address problems of privacy and fairness.
        \item While the authors might fear that complete honesty about limitations might be used by reviewers as grounds for rejection, a worse outcome might be that reviewers discover limitations that aren't acknowledged in the paper. The authors should use their best judgment and recognize that individual actions in favor of transparency play an important role in developing norms that preserve the integrity of the community. Reviewers will be specifically instructed to not penalize honesty concerning limitations.
    \end{itemize}

\item {\bf Theory assumptions and proofs}
    \item[] Question: For each theoretical result, does the paper provide the full set of assumptions and a complete (and correct) proof?
    \item[] Answer: \answerNA{} % Replace by \answerYes{}, \answerNo{}, or \answerNA{}.
    \item[] Justification: The paper does not include theoretical results.
    \item[] Guidelines:
    \begin{itemize}
        \item The answer \answerNA{} means that the paper does not include theoretical results. 
        \item All the theorems, formulas, and proofs in the paper should be numbered and cross-referenced.
        \item All assumptions should be clearly stated or referenced in the statement of any theorems.
        \item The proofs can either appear in the main paper or the supplemental material, but if they appear in the supplemental material, the authors are encouraged to provide a short proof sketch to provide intuition. 
        \item Inversely, any informal proof provided in the core of the paper should be complemented by formal proofs provided in appendix or supplemental material.
        \item Theorems and Lemmas that the proof relies upon should be properly referenced. 
    \end{itemize}

    \item {\bf Experimental result reproducibility}
    \item[] Question: Does the paper fully disclose all the information needed to reproduce the main experimental results of the paper to the extent that it affects the main claims and/or conclusions of the paper (regardless of whether the code and data are provided or not)?
    \item[] Answer: \answerYes{} % Replace by \answerYes{}, \answerNo{}, or \answerNA{}.
    \item[] Justification: Section \ref{sec:experiments} and the appendix include all the information needed to reproduce the main experimental results of the paper.
    \item[] Guidelines:
    \begin{itemize}
        \item The answer \answerNA{} means that the paper does not include experiments.
        \item If the paper includes experiments, a \answerNo{} answer to this question will not be perceived well by the reviewers: Making the paper reproducible is important, regardless of whether the code and data are provided or not.
        \item If the contribution is a dataset and\slash or model, the authors should describe the steps taken to make their results reproducible or verifiable. 
        \item Depending on the contribution, reproducibility can be accomplished in various ways. For example, if the contribution is a novel architecture, describing the architecture fully might suffice, or if the contribution is a specific model and empirical evaluation, it may be necessary to either make it possible for others to replicate the model with the same dataset, or provide access to the model. In general. releasing code and data is often one good way to accomplish this, but reproducibility can also be provided via detailed instructions for how to replicate the results, access to a hosted model (e.g., in the case of a large language model), releasing of a model checkpoint, or other means that are appropriate to the research performed.
        \item While NeurIPS does not require releasing code, the conference does require all submissions to provide some reasonable avenue for reproducibility, which may depend on the nature of the contribution. For example
        \begin{enumerate}
            \item If the contribution is primarily a new algorithm, the paper should make it clear how to reproduce that algorithm.
            \item If the contribution is primarily a new model architecture, the paper should describe the architecture clearly and fully.
            \item If the contribution is a new model (e.g., a large language model), then there should either be a way to access this model for reproducing the results or a way to reproduce the model (e.g., with an open-source dataset or instructions for how to construct the dataset).
            \item We recognize that reproducibility may be tricky in some cases, in which case authors are welcome to describe the particular way they provide for reproducibility. In the case of closed-source models, it may be that access to the model is limited in some way (e.g., to registered users), but it should be possible for other researchers to have some path to reproducing or verifying the results.
        \end{enumerate}
    \end{itemize}

\item {\bf Open access to data and code}
    \item[] Question: Does the paper provide open access to the data and code, with sufficient instructions to faithfully reproduce the main experimental results, as described in supplemental material?
    \item[] Answer: \answerYes{} % Replace by \answerYes{}, \answerNo{}, or \answerNA{}.
    \item[] Justification: Code is available at \url{https://github.com/3525starmoon-glitch/NiSM}, and the data are available upon request (see the appendix).
    \item[] Guidelines:
    \begin{itemize}
        \item The answer \answerNA{} means that paper does not include experiments requiring code.
        \item Please see the NeurIPS code and data submission guidelines (\url{https://neurips.cc/public/guides/CodeSubmissionPolicy}) for more details.
        \item While we encourage the release of code and data, we understand that this might not be possible, so \answerNo{} is an acceptable answer. Papers cannot be rejected simply for not including code, unless this is central to the contribution (e.g., for a new open-source benchmark).
        \item The instructions should contain the exact command and environment needed to run to reproduce the results. See the NeurIPS code and data submission guidelines (\url{https://neurips.cc/public/guides/CodeSubmissionPolicy}) for more details.
        \item The authors should provide instructions on data access and preparation, including how to access the raw data, preprocessed data, intermediate data, and generated data, etc.
        \item The authors should provide scripts to reproduce all experimental results for the new proposed method and baselines. If only a subset of experiments are reproducible, they should state which ones are omitted from the script and why.
        \item At submission time, to preserve anonymity, the authors should release anonymized versions (if applicable).
        \item Providing as much information as possible in supplemental material (appended to the paper) is recommended, but including URLs to data and code is permitted.
    \end{itemize}

\item {\bf Experimental setting/details}
    \item[] Question: Does the paper specify all the training and test details (e.g., data splits, hyperparameters, how they were chosen, type of optimizer) necessary to understand the results?
    \item[] Answer: \answerYes{} % Replace by \answerYes{}, \answerNo{}, or \answerNA{}.
    \item[] Justification: Section \ref{sec:experiments} and the appendix include all the training and test details.
    \item[] Guidelines:
    \begin{itemize}
        \item The answer \answerNA{} means that the paper does not include experiments.
        \item The experimental setting should be presented in the core of the paper to a level of detail that is necessary to appreciate the results and make sense of them.
        \item The full details can be provided either with the code, in appendix, or as supplemental material.
    \end{itemize}

\item {\bf Experiment statistical significance}
    \item[] Question: Does the paper report error bars suitably and correctly defined or other appropriate information about the statistical significance of the experiments?
    \item[] Answer: \answerYes{} % Replace by \answerYes{}, \answerNo{}, or \answerNA{}.
    \item[] Justification: Table \ref{tab:performance_benchmark} includes the main quantitative result with error bars.
    \item[] Guidelines:
    \begin{itemize}
        \item The answer \answerNA{} means that the paper does not include experiments.
        \item The authors should answer \answerYes{} if the results are accompanied by error bars, confidence intervals, or statistical significance tests, at least for the experiments that support the main claims of the paper.
        \item The factors of variability that the error bars are capturing should be clearly stated (for example, train/test split, initialization, random drawing of some parameter, or overall run with given experimental conditions).
        \item The method for calculating the error bars should be explained (closed form formula, call to a library function, bootstrap, etc.)
        \item The assumptions made should be given (e.g., Normally distributed errors).
        \item It should be clear whether the error bar is the standard deviation or the standard error of the mean.
        \item It is OK to report 1-sigma error bars, but one should state it. The authors should preferably report a 2-sigma error bar than state that they have a 96\% CI, if the hypothesis of Normality of errors is not verified.
        \item For asymmetric distributions, the authors should be careful not to show in tables or figures symmetric error bars that would yield results that are out of range (e.g., negative error rates).
        \item If error bars are reported in tables or plots, the authors should explain in the text how they were calculated and reference the corresponding figures or tables in the text.
    \end{itemize}

\item {\bf Experiments compute resources}
    \item[] Question: For each experiment, does the paper provide sufficient information on the computer resources (type of compute workers, memory, time of execution) needed to reproduce the experiments?
    \item[] Answer: \answerYes{} % Replace by \answerYes{}, \answerNo{}, or \answerNA{}.
    \item[] Justification: The appendix includes information on the computer resources.
    \item[] Guidelines:
    \begin{itemize}
        \item The answer \answerNA{} means that the paper does not include experiments.
        \item The paper should indicate the type of compute workers CPU or GPU, internal cluster, or cloud provider, including relevant memory and storage.
        \item The paper should provide the amount of compute required for each of the individual experimental runs as well as estimate the total compute. 
        \item The paper should disclose whether the full research project required more compute than the experiments reported in the paper (e.g., preliminary or failed experiments that didn't make it into the paper). 
    \end{itemize}
    
\item {\bf Code of ethics}
    \item[] Question: Does the research conducted in the paper conform, in every respect, with the NeurIPS Code of Ethics \url{https://neurips.cc/public/EthicsGuidelines}?
    \item[] Answer: \answerYes{} % Replace by \answerYes{}, \answerNo{}, or \answerNA{}.
    \item[] Justification: We reviewed NeurIPS Code Of Ethics.
    \item[] Guidelines:
    \begin{itemize}
        \item The answer \answerNA{} means that the authors have not reviewed the NeurIPS Code of Ethics.
        \item If the authors answer \answerNo, they should explain the special circumstances that require a deviation from the Code of Ethics.
        \item The authors should make sure to preserve anonymity (e.g., if there is a special consideration due to laws or regulations in their jurisdiction).
    \end{itemize}

\item {\bf Broader impacts}
    \item[] Question: Does the paper discuss both potential positive societal impacts and negative societal impacts of the work performed?
    \item[] Answer: \answerYes{} % Replace by \answerYes{}, \answerNo{}, or \answerNA{}.
    \item[] Justification: The appendix includes societal impacts.
    \item[] Guidelines:
    \begin{itemize}
        \item The answer \answerNA{} means that there is no societal impact of the work performed.
        \item If the authors answer \answerNA{} or \answerNo, they should explain why their work has no societal impact or why the paper does not address societal impact.
        \item Examples of negative societal impacts include potential malicious or unintended uses (e.g., disinformation, generating fake profiles, surveillance), fairness considerations (e.g., deployment of technologies that could make decisions that unfairly impact specific groups), privacy considerations, and security considerations.
        \item The conference expects that many papers will be foundational research and not tied to particular applications, let alone deployments. However, if there is a direct path to any negative applications, the authors should point it out. For example, it is legitimate to point out that an improvement in the quality of generative models could be used to generate Deepfakes for disinformation. On the other hand, it is not needed to point out that a generic algorithm for optimizing neural networks could enable people to train models that generate Deepfakes faster.
        \item The authors should consider possible harms that could arise when the technology is being used as intended and functioning correctly, harms that could arise when the technology is being used as intended but gives incorrect results, and harms following from (intentional or unintentional) misuse of the technology.
        \item If there are negative societal impacts, the authors could also discuss possible mitigation strategies (e.g., gated release of models, providing defenses in addition to attacks, mechanisms for monitoring misuse, mechanisms to monitor how a system learns from feedback over time, improving the efficiency and accessibility of ML).
    \end{itemize}
    
\item {\bf Safeguards}
    \item[] Question: Does the paper describe safeguards that have been put in place for responsible release of data or models that have a high risk for misuse (e.g., pre-trained language models, image generators, or scraped datasets)?
    \item[] Answer: \answerNA{} % Replace by \answerYes{}, \answerNo{}, or \answerNA{}.
    \item[] Justification: This paper does not include data or models that have a high risk for misuse.
    \item[] Guidelines:
    \begin{itemize}
        \item The answer \answerNA{} means that the paper poses no such risks.
        \item Released models that have a high risk for misuse or dual-use should be released with necessary safeguards to allow for controlled use of the model, for example by requiring that users adhere to usage guidelines or restrictions to access the model or implementing safety filters. 
        \item Datasets that have been scraped from the Internet could pose safety risks. The authors should describe how they avoided releasing unsafe images.
        \item We recognize that providing effective safeguards is challenging, and many papers do not require this, but we encourage authors to take this into account and make a best faith effort.
    \end{itemize}

\item {\bf Licenses for existing assets}
    \item[] Question: Are the creators or original owners of assets (e.g., code, data, models), used in the paper, properly credited and are the license and terms of use explicitly mentioned and properly respected?
    \item[] Answer: \answerYes{} % Replace by \answerYes{}, \answerNo{}, or \answerNA{}.
    \item[] Justification: We checked the licenses of assets.
    \item[] Guidelines:
    \begin{itemize}
        \item The answer \answerNA{} means that the paper does not use existing assets.
        \item The authors should cite the original paper that produced the code package or dataset.
        \item The authors should state which version of the asset is used and, if possible, include a URL.
        \item The name of the license (e.g., CC-BY 4.0) should be included for each asset.
        \item For scraped data from a particular source (e.g., website), the copyright and terms of service of that source should be provided.
        \item If assets are released, the license, copyright information, and terms of use in the package should be provided. For popular datasets, \url{paperswithcode.com/datasets} has curated licenses for some datasets. Their licensing guide can help determine the license of a dataset.
        \item For existing datasets that are re-packaged, both the original license and the license of the derived asset (if it has changed) should be provided.
        \item If this information is not available online, the authors are encouraged to reach out to the asset's creators.
    \end{itemize}

\item {\bf New assets}
    \item[] Question: Are new assets introduced in the paper well documented and is the documentation provided alongside the assets?
    \item[] Answer: \answerYes{} % Replace by \answerYes{}, \answerNo{}, or \answerNA{}.
    \item[] Justification: The licenses and data-use terms of the released code and data are stated in the appendix.
    \item[] Guidelines:
    \begin{itemize}
        \item The answer \answerNA{} means that the paper does not release new assets.
        \item Researchers should communicate the details of the dataset\slash code\slash model as part of their submissions via structured templates. This includes details about training, license, limitations, etc. 
        \item The paper should discuss whether and how consent was obtained from people whose asset is used.
        \item At submission time, remember to anonymize your assets (if applicable). You can either create an anonymized URL or include an anonymized zip file.
    \end{itemize}

\item {\bf Crowdsourcing and research with human subjects}
    \item[] Question: For crowdsourcing experiments and research with human subjects, does the paper include the full text of instructions given to participants and screenshots, if applicable, as well as details about compensation (if any)? 
    \item[] Answer: \answerYes{} % Replace by \answerYes{}, \answerNo{}, or \answerNA{}.
    \item[] Justification: Section \ref{sec:self_collected_data} and the appendix describe the study procedure, the tasks performed by participants, and the informed-consent process.
    \item[] Guidelines:
    \begin{itemize}
        \item The answer \answerNA{} means that the paper does not involve crowdsourcing nor research with human subjects.
        \item Including this information in the supplemental material is fine, but if the main contribution of the paper involves human subjects, then as much detail as possible should be included in the main paper. 
        \item According to the NeurIPS Code of Ethics, workers involved in data collection, curation, or other labor should be paid at least the minimum wage in the country of the data collector. 
    \end{itemize}

\item {\bf Institutional review board (IRB) approvals or equivalent for research with human subjects}
    \item[] Question: Does the paper describe potential risks incurred by study participants, whether such risks were disclosed to the subjects, and whether Institutional Review Board (IRB) approvals (or an equivalent approval/review based on the requirements of your country or institution) were obtained?
    \item[] Answer: \answerYes{} % Replace by \answerYes{}, \answerNo{}, or \answerNA{}.
    \item[] Justification: Section \ref{sec:self_collected_data} and the appendix includes the text on IRB approvals. Human study in this work does not give rise to clinical risks.
    \item[] Guidelines:
    \begin{itemize}
        \item The answer \answerNA{} means that the paper does not involve crowdsourcing nor research with human subjects.
        \item Depending on the country in which research is conducted, IRB approval (or equivalent) may be required for any human subjects research. If you obtained IRB approval, you should clearly state this in the paper. 
        \item We recognize that the procedures for this may vary significantly between institutions and locations, and we expect authors to adhere to the NeurIPS Code of Ethics and the guidelines for their institution. 
        \item For initial submissions, do not include any information that would break anonymity (if applicable), such as the institution conducting the review.
    \end{itemize}

\item {\bf Declaration of LLM usage}
    \item[] Question: Does the paper describe the usage of LLMs if it is an important, original, or non-standard component of the core methods in this research? Note that if the LLM is used only for writing, editing, or formatting purposes and does \emph{not} impact the core methodology, scientific rigor, or originality of the research, declaration is not required.
    %this research? 
    \item[] Answer: \answerNA{} % Replace by \answerYes{}, \answerNo{}, or \answerNA{}.
    \item[] Justification: LLM is used only for editing and formatting purposes. 
    \item[] Guidelines:
    \begin{itemize}
        \item The answer \answerNA{} means that the core method development in this research does not involve LLMs as any important, original, or non-standard components.
        \item Please refer to our LLM policy in the NeurIPS handbook for what should or should not be described.
    \end{itemize}

\end{enumerate}

\end{document}